\documentclass{article}

\usepackage{microtype}
\usepackage{graphicx}    
\usepackage{subcaption}  
\usepackage{booktabs} 

\usepackage{hyperref}
\usepackage{color}

\usepackage[preprint]{icml2026}

\usepackage{amsmath}
\usepackage{amssymb}
\usepackage{mathtools}
\usepackage{amsthm}
\usepackage{mathrsfs}
\usepackage{dsfont}

\usepackage{tikz}
\usetikzlibrary{arrows.meta,positioning}

\usepackage[capitalize,noabbrev]{cleveref}

\theoremstyle{plain}
\newtheorem{theorem}{Theorem}[section]
\newtheorem{proposition}[theorem]{Proposition}

\theoremstyle{definition}

\theoremstyle{remark}

\def \R{\mathbb{R}}

\def \E{\mathbb{E}}

\def \P{\mathbb{P}}
\def \Q{\mathbb{Q}}
\def \S{\mathbb{S}}

\def \W{\mathbb{W}}

\def \a{\alpha}
\def \b{\beta}
\def \s{\sigma}
\def \t{\theta}

\def \eps{\varepsilon}

\def\beqs{\begin{eqnarray*}}
\def\enqs{\end{eqnarray*}}
\def\beq{\begin{equation}}
\def\enq{\end{equation}}

\DeclareMathOperator*{\argmin}{argmin}

\def \Inf{\displaystyle\inf}
\def \Sup{\displaystyle\sup}

\def \Pc{{\cal P}}
\def \Hc{{\cal H}}

\def \msX{\mathscr{X}} 
\def \msY{\mathscr{Y}} 
\def \d{\mathrm{d}}

\def \Fc{{\cal F}}

\def \Nc{{\cal N}}

\def \Sc{{\cal S}}
\def \Tc{{\cal T}}

\def \Zc{{\cal Z}}

\def \mra{\mathrm{a}}

\usepackage[textsize=tiny]{todonotes}

\icmltitlerunning{LightSBB-M: Bridging Schr\"odinger and Bass for Generative Diffusion Modeling}

\begin{document}

\twocolumn[
\icmltitle{LightSBB-M: Bridging Schr\"odinger and Bass for \\
Generative Diffusion Modeling}



\icmlsetsymbol{equal}{*}

\begin{icmlauthorlist}
\icmlauthor{Alexandre ALOUADI}{bnpx}
\icmlauthor{Pierre HENRY-LABORDERE}{qube}
\icmlauthor{Grégoire LOEPER}{bnp}
\icmlauthor{Othmane MAZHAR}{upc}
\icmlauthor{Huyên PHAM}{x}
\icmlauthor{Nizar TOUZI}{nyu}
\end{icmlauthorlist}

\icmlaffiliation{bnpx}{École Polytechnique \& BNP PARIBAS}
\icmlaffiliation{bnp}{BNP PARIBAS}
\icmlaffiliation{upc}{Université Paris Cité}
\icmlaffiliation{x}{École Polytechnique}
\icmlaffiliation{qube}{Qube RT}
\icmlaffiliation{nyu}{New York University}

\icmlcorrespondingauthor{Alexandre ALOUADI}{alexandre.alouadi@bnpparibas.com}
\icmlcorrespondingauthor{Huyên PHAM}{huyen.pham@polytechnique.edu}

\icmlkeywords{Machine Learning, ICML}

\vskip 0.3in
]



\printAffiliationsAndNotice{}  

\begin{abstract}
The Schr\"odinger Bridge and Bass (SBB) formulation, which jointly controls drift and volatility, is an established extension of the classical Schr\"odinger Bridge (SB).  Building on this framework, we introduce \textbf{LightSBB‑M}, an algorithm that computes the optimal SBB transport plan in only a few iterations. The method exploits a dual representation of the SBB objective to obtain analytic expressions for the optimal drift and volatility, and it incorporates a tunable parameter $\beta>0$ that interpolates between pure drift (the Schr\"odinger Bridge) and pure volatility (Bass martingale transport). We show that LightSBB‑M achieves the lowest 2‑Wasserstein distance on synthetic datasets against state‑of‑the‑art SB and diffusion
baselines with an average of 19\% improvement. We also illustrate the generative capability of the framework on an unpaired image‑to‑image translation task (\textit{adult} $\rightarrow$ \textit{child} faces in FFHQ). These findings demonstrate that \textbf{LightSBB‑M} provides a scalable, high‑fidelity SBB solver that outperforms existing SB and diffusion baselines across both synthetic and real‑world generative tasks. The code is available at \href{https://github.com/alexouadi/LightSBB-M}{https://github.com/alexouadi/LightSBB-M}.
\end{abstract}

\section{Introduction}

Stochastic-process-based generative modeling has recently gained significant attention, notably through diffusion models and Schr\"odinger Bridge (SB) formulations, which construct a controlled stochastic dynamics transporting a source distribution $\mu_0$ to a target distribution $\mu_T$ \citep{Schrodinger1932, leonard2014survey, de2021diffusion}. By minimizing the relative entropy with respect to a Brownian prior, SB establishes deep connections between optimal transport, stochastic control, and score-based learning. However, the classical SB framework relies on a fixed volatility and requires finite entropy with respect to the Wiener measure, which limits its applicability for heavy-tailed or singular target distributions.

Bass martingale transport provides an alternative formulation by optimizing over volatility while enforcing a zero drift constraint \citep{bass1983skorokhod, backhoff2025geometric}. While this approach allows for greater flexibility on the marginals, it departs from diffusion-based generative mechanisms and lacks an explicit drift control. The Schr\"odinger-Bass Bridge (SBB) problem \cite{henrylabordere2026bridgingschrodingerbasssemimartingale} unifies these two paradigms by jointly optimizing over drift and volatility, introducing a parameter $\beta > 0$ that interpolates between the classical SB limit ($\beta \to \infty$) and the Bass martingale transport regime ($\beta \to 0$). A key practical motivation for SBB is that, unlike classical SB under a Brownian prior, it is not tied to the finite-entropy requirement $\mathrm{KL}(\mathbb{P}\|\mathbb{W}^\varepsilon) < \infty$. This matters already for simple heavy-tailed targets: when $\mu_0 = \delta_0$ and $\mu_T = \mathcal{T}(2)$, the corresponding entropy term diverges since $x^2 p_{\mathcal{T}(2)}(x) \sim 1/|x|$ for large $|x|$, so classical SB is infeasible, whereas the SBB objective remains well posed. This perspective is closely related to the recent literature on heavy-tailed generative modeling, where Gaussian assumptions are replaced by heavy-tailed alternatives such as $\alpha$-stable noise or Student-$t$ perturbation kernels to improve tail coverage \citep{shariatian2025, pandey2025}. In contrast, our approach is complementary: instead of changing the noise family itself, SBB retains a constant-volatility bridge in the transformed $Y$-space and gains flexibility through a learned transport-induced state-dependent volatility in the original space.

Despite its theoretical appeal, the SBB problem has received limited attention from a computational perspective, especially in high-dimensional generative modeling. Extending existing SB solvers to this setting is challenging due to the presence of stochastic volatility and implicit transport maps.

In this work, we introduce \textbf{LightSBB-M}, a simulation-efficient algorithm for solving the SBB problem. Leveraging a dual formulation of SBB and recent advances in Light Schr\"odinger Bridge matching \citep{pmlr-v235-gushchin24a}, our approach yields analytic expressions for the optimal drift and volatility and converges in only a few iterations.

Our main contributions are summarized as follows:
\begin{itemize}
    \item We propose a practical and scalable algorithm to compute the optimal Schr\"odinger-Bass Bridge, enabling efficient generative modeling with jointly learned drift and volatility.
    \item We derive explicit and tractable expressions for the optimal SBB controls, allowing simulation-free sampling and avoiding costly SDE discretization.
    \item We provide extensive numerical experiments on synthetic and high-dimensional datasets, demonstrating improved transport accuracy and generative performance compared to state-of-the-art Schr\"odinger Bridge and diffusion-based baselines.
\end{itemize}

This work advances the interface between optimal transport theory and modern generative modeling by making stochastic-volatility transport computationally tractable at scale. Beyond images, the framework opens new directions for generative modeling under distributional constraints, heavy-tailed data, and time-series with heteroskedasticity—settings that remain challenging for standard diffusion models.

\section{Background}
\subsection{The Schr\"odinger Bridge Problem}

In this section, we recall the formulation of the Schr\"odinger Bridge Problem (SBP) \cite{Schrodinger1932}.  
Let $\Omega = \mathcal{C}([0, T], \R^d)$ denote the space of continuous $\R^d$-valued paths on $[0, T]$, with $T < \infty$.  
Let $X = (X_t)_{t \in [0,T]}$ be the canonical process, and let $\mathcal{P}(\Omega)$ denote the set of probability measures on the path space $\Omega$. 
For $\P$ $\in$ $\mathcal{P}(\Omega)$, $\P_t$ $=$ $X_t\#\P$ is the marginal law of $X_t$ under $\P$ at time $t$. 

Let $\mu_0$ and $\mu_T$ be two probability distributions on $\R^d$. 
Let $\W^\eps$ denote the Wiener measure with variance $\eps$ $>$ $0$ on $\Omega$, which represents the prior belief about the system dynamics before observing data.  
The SBP is then formulated as the following entropy-minimization problem:
\begin{equation}
\label{eq:sbp_problem_kl}
\P^*  \in \argmin_{\mathbb{P}} \big\{ \mathrm{KL}(\mathbb{P}\|\mathbb{W}^\eps) : \mathbb{P} \in \mathcal{P}(\Omega), \, \mathbb{P}_0 = \mu_0, \, \mathbb{P}_T = \mu_T \big\},
\end{equation}
where $\mathrm{KL}(\mathbb{P}\|\mathbb{W}^\eps) = \E_{\P} \big[\ln \tfrac{d\P}{d\W^\eps} \big]$ denotes the Kullback–Leibler (KL) divergence between $\P$ and $\W$.

If $\mathrm{KL}(\mathbb{P}\|\mathbb{W}^\eps) < \infty$, then by Girsanov's theorem, the optimization problem \eqref{eq:sbp_problem_kl} admits a stochastic control formulation.  
Specifically, one seeks an $\R^d$-valued control process $\alpha$ satisfying $\E_\P \big[\int_0^T \|\alpha_t\|^2 \, \d t \big] < \infty$ such that
\begin{equation}
\label{eq:sbp_pb_control}
\alpha^*  \in \argmin_{\alpha} \Big\{ \E_\P \big[\int_0^T \|\alpha_t\|^2 \, \d t \big] :  X_0 \sim \mu_0, \, X_T \sim \mu_T \Big\},
\end{equation}
with the controlled dynamics $\d X_t = \alpha_t \, \d t + \sqrt{\eps} \, \d W_t, \; W \text{a Brownian motion under } \P$.

The associated path measure $\P^{SB} \in \mathcal{P}(\Omega)$ is the one closest to the Wiener path measure $\W^\eps$ in terms of KL divergence, and disintegrated as $\P^{SB} = \pi_{0,T}^{SB} \W^\eps_{|0,T}$, where $\pi_{0,T}^{SB}$ denotes the optimal coupling of the following static Entropic Optimal Transport (EOT) problem:
\begin{footnotesize}
\begin{equation*}
 \argmin_{\pi \in \Pc(\R^d\times\R^d)} 
 \Big\{ \E_\pi \big[| X_0 - X_T |^2\big] - 2 T \eps \, \Hc(\pi) :  \pi_0 = \mu_0, \, \pi_T = \mu_T \Big\},
\end{equation*}
\end{footnotesize} 
where $\Hc(\pi)$ denotes the entropy, and $\W^\eps_{|0,T}$ is the law of the Brownian bridge between times $0$ and $T$. The optimal EOT plan takes the form $\pi_{0,T}^{SB}(x_0,x_T) =  \nu_0(x_0) \,
e^{-\frac{|x_0 - x_T|^2}{2\eps T}} \,
e^{\psi(x_T)}$
where the so-called \emph{potentials} $\nu_0$ and $\psi = \log h_T$ satisfy the Schr\"odinger system under the marginal constraints:
\begin{equation} 
\label{eq:schrodinger-system}
	\begin{cases} \mu_T \;=\; h_T \nu_T, \quad \nu_T(x) \;=\;  \nu_0*\Nc_{\eps T}(x), \\
		 \mu_0 \;=\; h_0 \nu_0, \quad h_0(x) \;=\;  h_T *\Nc_{\eps T}(x).
	  \end{cases} 
\end{equation}
where $\Nc_{\s^2}$ denotes the Gaussian distribution with variance $\s^2$. Finally, the optimal drift is given by the score function $\alpha^*(t,x) =  \eps \nabla_x \log h_t(x),  \; (t,x) \in [0,T) \times \R^d$ where $h_t = h_T * \Nc_{\eps(T-t)}$.

\subsection{Diffusion Schr\"odinger Bridge for Generative Modeling}

Given samples from $\mu_0$ and $\mu_T$, the objective is to learn the solution of the SBP in order to generate new samples from $\mu_T$ given unseen samples from $\mu_0$. To achieve this, one may either learn the path measure $\P^{SB} = \pi^{SB}_{0,T}\W^{\eps}_{|0,T}$ solving \eqref{eq:sbp_problem_kl}, where the Brownian bridge component $\W^{\eps}_{|0,T}$ is known and only the optimal coupling $\pi^{SB}_{0,T}$ needs to be estimated, or alternatively learn the optimal drift $\a^*$ solving \eqref{eq:sbp_pb_control}, and then simulate the controlled dynamics $\d X_t = \a_t^* \d t + \sqrt{\eps}\d W_t$, starting from $X_0 \sim \mu_0$, to obtain new samples $X_T \sim \mu_T$. Several recent and competitive \emph{Diffusion Schr\"odinger Bridge} (DSB) solvers have been proposed to address this learning problem.


\subsubsection{Sinkhorn Algorithm}

Recall the optimal plan admits the separable form
\[
\pi^{\mathrm{SB}}_{0,T}(x_0,x_T)
=\nu_0(x_0)\,e^{-\frac{|x_0-x_T|^2}{2\varepsilon T}}\,e^{\psi(x_T)},
\]
where the Schrödinger potentials $(\nu_0,\psi)$ solve the Schrödinger system \eqref{eq:schrodinger-system}, with $\psi = \log h_T$.

The Sinkhorn algorithm computes $(\nu_0,h_T)$ by iteratively enforcing the marginal constraints, see e.g.\ \cite{leonard2014survey,PeyreCuturi2019}. Starting from an initial guess $h_T^{(0)}$, the iterations read
\[
\begin{cases}
\nu_0^{(k+1)}(x)
= \dfrac{\mu_0(x)}{h_0^{(k)}(x)},
\quad h_0^{(k)} = h_T^{(k)} \ast \mathcal N_{\varepsilon T},\\[0.4em]
h_T^{(k+1)}(x)
= \dfrac{\mu_T(x)}{\nu_T^{(k+1)}(x)},
\quad \nu_T^{(k+1)} = \nu_0^{(k+1)} \ast \mathcal N_{\varepsilon T}.
\end{cases}
\]
At convergence, one recovers the SB coupling $\pi^{\mathrm{SB}}_{0,T}$ and the associated SB drift
$\alpha^\ast(t,x)=\varepsilon\nabla_x\log h_t(x)$ with
$h_t = h_T \ast \mathcal N_{\varepsilon(T-t)}$, see \cite{Follmer1988,leonard2014survey}.

While Sinkhorn is simulation-free and easy to implement, it suffers from important drawbacks in practice. Its reliance on repeated Gaussian convolutions makes it poorly scalable in high dimension and numerically unstable in the small-noise regime $\varepsilon\to0$ \cite{PeyreCuturi2019,de2021diffusion}. Moreover, Sinkhorn only provides the endpoint coupling and Schrödinger potentials, and does not yield a parametric or dynamic representation of the SB process, which limits its applicability to generative modeling \cite{NEURIPS2023_c428adf7}.

\subsubsection{Iterative Markovian Fitting}

We recall that $\P^{SB}$ is the unique measure satisfying the boundary conditions $\P^{SB}_0 = \pi_0$ and $\P^{SB}_T = \pi_T$ such that $\P^{SB}$ is Markovian and $\P^{SB}_{|0,T} = \W^\eps_{|0,T}$ simultaneously \cite{leonard2014survey}. If the last condition is verified, we say that $\P^{SB} \in \cal{R(\W^\eps)}$ the reciprocal class of the reference measure $\mathbb{W}^\eps$ \cite{NEURIPS2023_c428adf7}. In other words, given an initial point $x_0$ and a terminal point $x_T$, the law of the process $X$ is a Brownian motion. The \emph{Iterative Markovian Fitting} (IMF) algorithm \cite{NEURIPS2023_c428adf7} provides an alternative methodology for computing $\P^{SB}$. It constructs a sequence of path measures $(\mathbb{P}^n)_{n\in\mathbb{N}}$ by alternating between two projections:
\beqs
 \begin{cases} 
    \mathbb{P}^{2n+1} &= \operatorname{proj}_{\mathcal{M}}(\mathbb{P}^{2n}), \\
    \mathbb{P}^{2n+2} &= \operatorname{proj}_{\mathcal{R}(\mathbb{W}^\eps)}(\mathbb{P}^{2n+1}),
\end{cases}
\enqs
initialized with a measure $\mathbb{P}^0 \in \mathcal{R}(\mathbb{W}^\eps)$ satisfying the boundary conditions $\mathbb{P}^0_0 = \pi_0$ and $\mathbb{P}^0_T = \pi_T$. Here, $\mathcal{M}$ denotes the space of Markov measures. The IMF iterates preserve the marginal constraints $\mathbb{P}^n_0 = \pi_0$ and $\mathbb{P}^n_T = \pi_T$ for all $n$ and converges to the unique SB, $\mathbb{P}^* = \mathbb{P}^{\mathrm{SB}}$. However, this method is not simulation-free and can suffer from error accumulation across iterations if the Markovian projection step is not learned perfectly.

\subsubsection{LightSB-M: Light Schr\"odinger Bridge Matching}
\label{sec:light_sbm}

The LightSB-M \cite{pmlr-v235-gushchin24a} method solves the SB problem via a single, \emph{optimal projection} step based on a new characterization of SB. Let $\pi$ $\in$ $\Pi(\mu_0,\mu_T)$ with $\Pi(\mu_0,\mu_T)$ the set of all transport plans between $\mu_0$ and $\mu_T$, and $\P^\pi$ $=$ $\pi \W^\eps_{|0,T}$  its reciprocal process.  Then \cite{pmlr-v235-gushchin24a}
\beq \label{eq:light_sb_kl}
\P^{SB} = \argmin_{\Q \in \Sc(\mu_0)} {\rm KL}(\P^\pi|\Q),  
\enq
where $\Sc(\mu_0)$ is the set of SB processes  starting from $\mu_0$. From the separable form of optimal EOT, the optimal plan associated to a process in  $\Sc(\mu_0)$ can be written in the disintegrated  form: 
 \beq
 \label{eq:plan_sb}
 \pi_{\varphi}(x_0,x_T)  =  \mu_0(x_0)  \underbrace{ \frac{ e^{<x_0,x_T>/\eps} \varphi(x_T)}{c_\varphi(x_0)}}_{\pi_\varphi(x_T|x_0)}, 
 \enq
where $c_\varphi(x_0)$ $:=$ $\int e^{<x_0,x_T>/\eps} \varphi(x_T) \d x_T$, and $\varphi$ is the adjusted potential. Setting $\Q_\varphi$ $:=$ $\pi_\varphi \W^\eps_{|0,T}$ $\in$ $\Sc(\mu_0)$ yields a tractable objective for \eqref{eq:light_sb_kl} \cite{pmlr-v235-gushchin24a}

 \beqs
 {\rm KL}(\P^\pi | \Q_\varphi) =  \frac{1}{2} \underbrace{\E^{\P^\pi} \Big[  \Big \| \a_\varphi(t,X_t) -  \frac{X_T - X_t}{T-t} \Big\|^2}_{{\rm DSM}(\varphi)} \Big]   + C(\pi), 
 \enqs
where $\a_\varphi$ is the score-drift of $\Q_\varphi$. We are then led to minimize over $\varphi$ the denoising score matching loss DSM$(\varphi)$ via stochastic gradient descent.  In practice, one parametrizes $\varphi$ by a mixture of Gaussian densities such that $\a_\varphi$ is analytic and requires no neural network \cite{korotin2024lightschrodingerbridge}. Once $\varphi^*$ is learnt,  we sample $X_0$ $\sim$ $\mu_0$, $X_T$ $\sim$ $\pi_{\varphi^*}(\cdot|X_0)$ $\sim$ $\mu_T$, and so we do not have to solve the associated SDE.

\section{Bridging Schr\"odinger and Bass}

The Schr\"odinger--Bass (SBB) problem is an extension of the classical SB problem by jointly optimizing over both drift and volatility. It is introduced and studied in \cite{henrylabordere2026bridgingschrodingerbasssemimartingale}.  
Given two distributions $\mu_0, \mu_T \in \Pc(\R^d)$, the goal is to minimize, over 
$\P \in \Pc(\mu_0,\mu_T) = \big\{ \P \in \Pc(\Omega) : X_0 \overset{\P}{\sim} \mu_0, \; X_T \overset{\P}{\sim} \mu_T \big\}$, the quadratic cost:  
\begin{equation}
\label{eq:sbb_quad_cost}
  J(\P) = \E_\P \Big[ \int_0^T \underbrace{\| \alpha_t \|^2 + \beta \| \sigma_t - \sqrt{\varepsilon} I_d \|^2}_{H_\beta(\alpha_t,\sigma_t)} \, \d t \Big],
\end{equation}
for some $\beta > 0$, where $(\alpha, \sigma)$ is the drift/volatility of $X$ under $\P$, i.e.,   $\d X_t = \alpha_t \, \d t + \sigma_t \, \d W_t$. 
This problem is denoted by 
${\rm SBB}(\mu_0,\mu_T) := \inf_{\P \in \Pc(\mu_0,\mu_T)} J(\P)$.

Note that as $\beta \to \infty$, the volatility $\sigma$ is constrained to equal $\sqrt{\varepsilon} I_d$, recovering the classical SB problem. Conversely, dividing \eqref{eq:sbb_quad_cost} by $\beta$ and letting $\beta \to 0$ forces the drift term $\alpha$ to vanish, yielding the Bass martingale transport problem. In other words, the parameter $\beta$ controls the relative weight of drift versus volatility, interpolating between these two well-known cases. We emphasize that the $\beta \downarrow 0$ limit is understood only as a formal interpolation in the present paper: a genuine zero-drift limit requires $(\mu_0, \mu_T)$ to satisfy convex order, and we do not prove that SBB optimizers converge to the specific Bass optimizer among all martingale couplings. The condition $\beta T > 1$ is required for dual attainment and is stated explicitly in  \citep[ Theorem~3.1(c)]{henrylabordere2026bridgingschrodingerbasssemimartingale}

\subsection{Dual Representation of the Primal SBB}

The primal problem ${\rm SBB}(\mu_0,\mu_T)$ admits a dual representation, which consists of maximizing over a suitable class of functions $(v,\psi)$ the Lagrangian functional
\beqs
L_{\mu_0, \mu_T}(\psi, v) = \int \psi(x) \, \mu_T(\d x)  -  \int v(0,x) \, \mu_0(\d x),
\enqs
where $v$ is the value function of the unconstrained stochastic control problem with Bellman equation: 
\beq
\label{eq:bellman_sbb}
\begin{cases}
\partial_t v + H_\beta^*( \nabla_x v, D_x^2 v) = 0, \quad  \text{on } [0,T) \times \R^d, \\
v(T,\cdot) = \psi,  \quad  \text{on }  \R^d,
\end{cases}
\enq
with $H_\beta^*$ denoting the Fenchel--Legendre transform of $H_\beta$, explicitly given by
\beqs
H_\beta^*(p,q) = \frac{1}{2}|p|^2 + \frac{\varepsilon \beta}{2} I_d : \Big( \big( I_d - \frac{q}{\beta} \big)^{-1} -  I_d \Big),
\enqs
for $(p,q) \in \R^d \times \S_+^d$ such that $q < \beta I_d$. Assuming that ${\rm SBB}(\mu_0,\mu_T) < \infty$, we have:

\begin{itemize}
\item {\bf Attainment of the primal problem}: there exists $(\alpha^*,\sigma^*) \leftrightarrow \P^{\rm SBB}$ attaining the infimum in ${\rm SBB}(\mu_0,\mu_T)$, in feedback form: $\alpha_t^* = \mra^*(t,X_t)$, $\sigma_t^* = \vartheta^*(t,X_t)$.  

\item {\bf Duality relation}: we have 
\begin{equation}   
\label{eq:dualproblem}
{\rm SBB}(\mu_0,\mu_T) = \sup_{\substack{
        \psi \in C^2 \cap C_b^\infty \cap L^1(\mu_T), \
        v \in C_b^{1,2}, \\
        D_x^2 v < \beta I_d, \
        \psi = v(T, \cdot),  \\
        \partial_t v + H_\beta^*(\nabla_x v, D_x^2 v) = 0 
    }}  L_{\mu_0, \mu_T}(\psi, v)
\end{equation}

\item {\bf Duality on the control}: 

When $\mu_0$ and $\mu_T$ have finite second moment, and if $\beta > \frac{1}{T}$, then the supremum in the dual problem is attained at $(v^*,\psi^*)$, and the optimal feedback policies are given by 
		\beqs
		\begin{cases}
			\mra^*(t,x) \; = \; \nabla_x v^*(t,x), \quad   (t,x) \in [0,T)\times\R^d,  \\
			\vartheta^*(t,x) \; = \; \sqrt{\eps} \Big(I_d - \frac{D_x^2 v^*(t, x)}{\beta} \Big)^{-1}.  
		\end{cases}
		\enqs
	\end{itemize}


\begin{proposition}[Envelope projection and admissible map]
\label{prop:envelope}
Assume $\beta T > 1$. For any $\phi \in C_w := \{ \psi \in C^0(\mathbb{R}^d, \mathbb{R}) : \left|\frac{\psi}{w}\right|_\infty < \infty \}$, define its $\beta$-convex envelope $\tilde\phi := \Tc_\beta^- \circ \Tc_\beta^+[\phi]$. Then:
\begin{enumerate}
    \item $\tilde\phi \in C_w^{\mathrm{conv}}$ and $J(\tilde\phi) \ge J(\phi)$;
    \item the associated exact value function $v_t^{\tilde\phi}(x) := \Tc_\beta^+[u_{T-t}^{\tilde\phi}](x)$ satisfies $D_x^2 v_t^{\tilde\phi}(x) < \beta I_d$ for $t < T$;
    \item the induced map $\msY_t^{\tilde\phi}(x) := x - \frac{1}{\beta} D v_t^{\tilde\phi}(x)$ is well-defined and one-to-one on $\R^d$ for every $t < T$, with $D\msY_t^{\tilde\phi}(x) > 0$.
\end{enumerate}
The proof follows from \citep[Remark~3.1, Lemma~5.1, Theorem~3.2(c)]{henrylabordere2026bridgingschrodingerbasssemimartingale} and is given in Appendix~\ref{app:proof_envelope}.
\end{proposition}

\subsection{SBB System}

One can exploit the quadratic form of the SBB criterion to reduce the dual problem to the maximization over the {\it potential} $\psi \in C^2 \cap C_b^\infty \cap L^1(\mu_T)$, subject to $D_x^2 \psi < \beta I_d$, of the Donsker--Varadhan type functional: 
\beqs 
\int \psi \d\mu_T 
 - \int \underbrace{\Tc_\beta^+\Big[ \varepsilon \log \int e^{\Tc_\beta^-[\psi](\cdot + z)}  \Nc_{\varepsilon T}(\d z) \Big]}_{v}(x) \mu_0(\d x), 
\enqs
where $\Tc_\beta^{\pm}$ are the {\bf quadratic inf/sup convolution} operators: 
\beqs
\begin{cases}
\Tc_\beta^+[\phi](x) := \Inf_{y \in \R^d} \big[ \phi(y) + \frac{\beta}{2} | x - y |^2 \big], \quad x \in \R^d, \\
\Tc_\beta^-[\psi](y) := \Sup_{x \in \R^d} \big[ \psi(x) - \frac{\beta}{2} | x - y |^2 \big], \quad y \in \R^d.  
\end{cases}
\enqs

	This leads to the SBB system, where a potential $\psi^*$ (or $\phi^* = \Tc_\beta^-[\psi^*] = \log h_T^*$) of the dual SBB problem satisfies: 
	\beqs
	\begin{cases}
	\msY_T \# \mu_T = h_T^* \ \nu_T, \quad \nu_T = \nu_0 * \Nc_{\varepsilon T} \\
	\msY_0 \# \mu_0 = h_0^* \, \nu_0, \quad h_0^* = h_T^* * \Nc_{\varepsilon T}
	\end{cases}
	\enqs
	where $\#$ is the pushforward operator, and the transport map $\msY_t$ is defined as : 
	\beqs  
	\msY_t = (\nabla_y \Phi_t)^{-1}, \quad \Phi_t(y) = \frac{|y|^2}{2} + \frac{1}{\beta} \eps \log h_t^*(y),
	\enqs
	which is an increasing (convex) function from $\R^d$ into $\R^d$ for any $t \in [0,T]$. Note that when $\beta \to \infty$, $\msY = \mathrm{I_d}$ and we recover the Schr\"odinger system. Conversely, when $\beta \to 0$, $h^*$ is constant and we recover the Bass system. The optimal drift and volatility of SBB are given by 
\beqs
\begin{cases}
\alpha_t^* = \varepsilon \nabla_y \log h_t^*(\msY_t(X_t)), \\  
\sigma_t^* = \sqrt{\varepsilon} D_y^2 \Phi_t(\msY_t(X_t)), 
\end{cases}
\quad t \in [0,T].
\enqs 

If we define the process 
\beq
\label{eq:msY_def}
Y_t = \msY_t(X_t) = X_t - \frac{1}{\beta} \eps \nabla_y \log h_t^*(\msY_t(X_t)), \; t \in [0,T],
\enq
and the change of measure $\frac{\d\Q^*}{\d\P^{\rm SBB}} \Big|_{\Fc_t} = \frac{1}{h_t^*(Y_t)}, \; t \in [0,T]$ (which is indeed a $\P^{\rm SBB}$-martingale with expectation $1$ by the SBB system), then
\begin{itemize}
\item $(Y_t)_t$ is a Brownian motion with volatility $\sqrt{\varepsilon}$ under $\Q^*$ with initial law $\nu_0$. 
\item $X_t = \msY_t^{-1}(Y_t) = Y_t + \frac{\varepsilon}{\beta} \nabla_y \log h_t^*(Y_t)$, $t \in [0,T]$, is a stretched Brownian motion under $\Q^*$.  
\item The dynamics under $\P^{\rm SBB}$ are:
\begin{equation*}
\begin{cases}
\d X_t = D_y^2 \Phi_t(Y_t) \, \d Y_t, \\
\d Y_t = \varepsilon \nabla_y \log h_t^*(Y_t) \, \d t + \sqrt{\varepsilon} \, \d W_t, 
\end{cases} 
\quad t \in [0,T].
\end{equation*}
\end{itemize}

In other words, $\P^{\rm SBB}$ is the Bass transport of a SB, i.e., a stretched SB. The SBB system can be visualized in the figure below.

\begin{tikzpicture}[
    >=Latex,
    font=\Large,
    scale=0.85,
    transform shape,
    box/.style={draw=none, fill=green!80!black, text=white, inner xsep=13pt, inner ysep=3pt},
    node distance=9mm
]

\node (muT) {\footnotesize{$\mu_T$}};
\node (yT)  [right=of muT] {\footnotesize{$\msY_T\#\mu_T$}};
\node (derT) [right=of yT, xshift=3mm] {\scriptsize{$\dfrac{d\,\msY_T\#\mu_T}{d \nu_T}=e^{{\phi^*}}$}};

\node (vT) [right=of derT, xshift=7mm] {\scriptsize{$\nu_T := {\cal N}_T * \nu_0$}};
\node (v0r) [below=20mm of vT] {\footnotesize{$\nu_0$}};

\node (mu0) [below=20mm of muT] {\footnotesize{$\mu_0$}};
\node (y0)  [right=of mu0] {\footnotesize{$\msY_0\#\mu_0$}};
\node (der0) [right=of y0, xshift=5mm] {\scriptsize{$\dfrac{d\,\msY_T\#\mu_T}{d\nu_0}={\cal N}_T * e^{{\phi^*}}$}};
\node (v0)  [right=of der0, xshift=5mm] {};

\draw[->, thick] (muT) -- (yT);
\draw[->, thick] (yT) -- (derT);
\draw[->, thick] (derT) -- (vT);

\draw[->, thick] (mu0) -- (y0);
\draw[->, thick] (y0) -- (der0);
\draw[->, thick] (der0) -- (v0);

\draw[->, thick] (mu0) -- (muT);

\draw[->, thick] (v0r) -- (vT);

\node[box] (bass) [below=7mm of yT] {Bass};
\node[box] (sch)  [below=5mm of derT] {Schr\"odinger};

\end{tikzpicture}


\section{Generative Modeling with SBB}

To generate new samples from $\mu_T$ via the learned 
SBB system,   one could directly simulate 
\begin{align*} 
\d X_t &= \;  \varepsilon \nabla_y \log h_t^*(\msY_t(X_t)) \, \d t \\
& \qquad + \sqrt{\varepsilon} \, D_y^2 \Phi_t(\msY_t(X_t)) \, \d W_t, \; X_0 \sim \mu_0,
\end{align*} 
using an SDE solver (e.g., the Euler--Maruyama scheme) to obtain $X_T \sim \mu_T$. However, this requires computing the inverse of a Hessian matrix, which can be challenging in high dimensions. To overcome this, one can instead generate the process $Y = \msY(X)$ as a DSB:
\beqs
\d Y_t = \varepsilon \nabla_y \log h_t^*(Y_t) \, \d t + \sqrt{\varepsilon} \, \d W_t,
\enqs
with $Y_0 \sim \msY_0 \# \mu_0$ and $Y_T \sim \msY_T \# \mu_T$, and score drift
\[
s_t^*(y)=\varepsilon \nabla_y \log h_t^*(y).
\]
Then, $X_T$ can be recovered via
\beqs
X_T = \msY_T^{-1}(Y_T) = Y_T + \frac{1}{\beta} s_T^*(Y_T)
\sim \mu_T.
\enqs

\subsection{Training}

As Section~\ref{sec:light_sbm} provides an efficient way to solve the SB between $\msY_0 \# \mu_0$ and $\msY_T \# \mu_T$, it remains to learn the transport map $\msY$. Exploiting that $\msY_t = \msX_t^{-1}$ with
\begin{equation}
    \msX_t(y) = y + \frac{1}{\beta} s_t^*(y),
\end{equation}
we learn the inverse of $\msX$ using a neural network $\Zc_{\tilde{\theta}}$ to obtain $\msY$, avoiding the need to solve the fixed point~\eqref{eq:msY_def}. Importantly, LightSBB-M does \emph{not} parameterize the SBB volatility $\sigma_t^*$ directly: the full matrix-valued volatility in $X$-space is recovered implicitly through the learned map $\msY_t$, while the transformed process $Y$ has constant volatility $\sqrt{\varepsilon} I_d$. This can be viewed as a multivariate, time-dependent Lamperti-type reparameterization: the change of variables converts a state-dependent diffusion in $X$-space into a constant-volatility process in $Y$-space, which succeeds because of the convex/gradient structure of the optimal map (Proposition~\ref{prop:envelope}). This is also computationally advantageous, since it avoids directly learning or evaluating a $d \times d$ volatility field. The numerically delicate regime is when the inverse map becomes poorly conditioned, in particular for very small $\beta$.

\textit{Implementation remark.} By Proposition~\ref{prop:envelope}, one may enforce the admissible-map structure at the dual level by replacing any candidate potential by its $\beta$-convex envelope. In additional experiments, we also tested structure-preserving neural architectures enforcing this monotonicity structure more directly. They did not yield a meaningful improvement in transport accuracy while increasing training time, so we retain the simpler MLP parameterization. This is consistent with recent stochastic-control and bridge-based numerical practice, where exact structural conditions are often handled at the PDE/control level while neural approximations remain unconstrained or softly constrained for scalability \citep{KimChoKimKim2025, HuaLauriereVandenEijnden2025, MaTanXu2025}; by contrast, some static OT methods enforce convexity explicitly through ICNN architectures when Brenier structure is the primary object \citep{MakkuvaTaghvaeiOhLee2020ICNNOT}.

We then introduce LightSBB-M (Algorithm~\ref{alg:lightsbb}) to efficiently solve the SBB problem as follows.

Let $\theta = \{\alpha_j, r_j, \Sigma_j\}_{j=1}^J$ denote the parameters of a Gaussian-mixture potential $v_\theta$, and $\tilde{\theta}$ the parameters of the neural network $Z_{\tilde{\theta}}$. Initialize $\phi^0 = 0$, hence $\msY^0 = \Zc_{\tilde{\theta}}^0 = I_d$. Then, for $0 \le k \le K$:

\begin{enumerate}
    \item \textbf{Endpoint sampling.} 
    Draw endpoint pairs $(y_0, y_T)$ from a coupling of $(p_0^k, p_T^k)$, where $p_t^k = \mathrm{d}(\msY_t^k \# \mu_t)/\mathrm{d}y$, using
\beqs
\begin{cases}
   \msY_0^k(x_0) = y_0 = \Zc_{\tilde{\theta}}^k(0, x_0), \quad x_0 \sim \mu_0, \\
   \msY_T^k(x_T) = y_T = \Zc_{\tilde{\theta}}^k(T, x_T), \quad x_T \sim \mu_T.
\end{cases}
\enqs

    \item \textbf{Bridge sampling.} 
    For $t \sim \mathcal{U}[0,T)$ and $Z \sim \mathcal{N}(0,I_d)$, sample $y_t \sim \W^{\varepsilon}_{|y_0, y_T}$ as an intermediate point from the Brownian bridge:
    \begin{equation}
    \label{eq:brownian_bridge}
        y_t = \frac{T - t}{T} y_0 + \frac{t}{T} y_T + \sigma_t \sqrt{\varepsilon} Z, \quad \sigma_t^2 = \frac{t(T-t)}{T}.
    \end{equation}

    \item \textbf{Regression step on $\theta$.} 
    Update $\theta^k$ by minimizing the bridge-matching loss
\begin{align}
\label{eq:loss_lightsb}
\mathcal{L}(\theta^{k})
  &= \E_{t \sim \mathcal{U}([0,T))} \,
     \E_{y_T \sim p_T,\, y_t \sim W^{\varepsilon}_{|y_0, y_T}} \nonumber\\
  &\quad \left[ \left\| s_\theta^{k}(t, y_t) - \frac{y_T - y_t}{T-t} \right\|_2^2 \right],
\end{align}
which is the KL projection onto the set of SBs, ensuring that the learned drift coincides with the true SB drift.

\item \textbf{Regression step on $\tilde{\theta}$.} Once $\theta^{k+1}$ is obtained, update $\tilde{\theta}^k$ by minimizing
\beq
\label{eq:loss_Z}
\begin{split}
\mathcal{L}(\tilde{\theta}^{k})
  &= \E_{x_0 \sim \mu_0,\, x_T \sim \mu_T} \Big[
      \| \Zc_{\tilde{\theta}}^{k}(0, \msX_0(x_0)) - x_0 \|^2 \\
  &\qquad\quad
      + \| \Zc_{\tilde{\theta}}^{k}(T, \msX_T(x_T)) - x_T \|^2
     \Big].
\end{split}
\enq
\end{enumerate}

Moreover, using a Gaussian-mixture parametrization of $v$ from LightSB-M, we can derive a closed-form expression for the drift $s_{\theta} \triangleq s_{v_{\theta}}$ \cite{pmlr-v235-gushchin24a} as follows:
\begin{equation}
\label{eq:drift_lightsbb}
\begin{split}
s_{\theta}(t, y)
  &= \varepsilon \, \nabla_y 
     \log \mathcal{N}\!\left(y \,\middle|\, 0,\, \varepsilon(T-t) I_d \right) \\
  &\quad \times 
     \sum_{j=1}^{J} \alpha_j\,
     \mathcal{N}\!\left(r_j \,\middle|\, 0,\, \varepsilon \Sigma_j \right)\,
     \mathcal{N}\!\left(h_j(t, y) \,\middle|\, 0,\, A^t_j \right),
\end{split}
\end{equation}
where $A^t_j \triangleq \frac{t}{\varepsilon(T-t)} I_d + \frac{1}{\varepsilon} \Sigma_j^{-1}$ and $h_j(t, y)\triangleq \frac{y}{\varepsilon(T-t)}+\frac{1} {\varepsilon} \Sigma_j^{-1} r_j$.

	\begin{algorithm}[tb]
		\caption{LightSBB-M Training Algorithm}\label{alg:lightsbb}
		\begin{algorithmic}
			 \STATE {\bfseries Input:} Samples $(x_0^m, x_T^m)_{m \leq M} \sim(\mu_0,\mu_T)$, $\theta = \{\alpha_j, \mu_j, \Sigma_j \}_{j \leq J}$, $\tilde{\theta}$, $\beta>0$, $K >0$ 
            
            \STATE \textbf{Initialization:} Start with $\msY^0 = I_d$ and $\Zc_{\tilde{\theta}}^0 = I_d$
			\FOR{$k=0, \cdots, K-1$}	
            
            \REPEAT
			\STATE Draw sample batch of pairs $(x_0^n, x_T^n)_{n \leq N}$
			\STATE Compute $\msY_0^k(x_0^n) =y_0^n = \Zc^k_{\tilde{\theta}}(0, x_0^n)$ and $\msY_T^k(x_T^n)=y_T^n = \Zc^k_{\tilde{\theta}}(T, x_T^n)$
            
            \STATE Sample batch $(y_t^n)_{n \leq N} \sim \mathbb{W}_{|y_0, y_1}$ using ~\eqref{eq:brownian_bridge}
            \STATE Compute the drift $s_{\theta}^k$ using ~\eqref{eq:drift_lightsbb} and update $\theta^{k}$ by minimizing ~\eqref{eq:loss_lightsb}
            \UNTIL{convergence}
            
            \STATE $\theta^{k+1} \leftarrow{\theta^k}$

            \REPEAT
			\STATE Draw sample batch of pairs $(x_0^n, x_T^n)_{n \leq N}$
            
			\STATE Compute $\msX_0(x_0^n) = x_0^n + \frac{1}{\beta} s_{\theta}^{k+1}(0, x_0^n)$ and $\msX_T(x_T^n) = x_T^n + \frac{1}{\beta} s_{\theta}^{k+1}(T, x_T^n)$
            
            \STATE Update $\tilde{\theta}^{k}$ by minimizing \eqref{eq:loss_Z}
            \UNTIL{convergence}
            \STATE $\tilde{\theta}^{k+1} \leftarrow{\tilde{\theta}^k}$

            \ENDFOR
			\STATE \textbf{Return} $\theta^K, \; \tilde{\theta}^{K}$ 
\end{algorithmic}
	\end{algorithm}
	
We also provide in Appendix~\ref{sec:other_algo} alternative algorithms, including a simplification of Algorithm~\ref{alg:lightsbb} when $\b$ is large and Sinkhorn-based solver for the SBB problem.

	Note that the regression loss \eqref{eq:loss_lightsb} rules out arbitrarily small $T$ as the target would explode. On the other hand, an excessively large $T$ drives the noisy marginal $\mu_T$ to become almost indistinguishable from $\mu_0$, forcing the reverse dynamics to undo an overwhelming amount of noise; this dramatically inflates the variance of the optimal control and deteriorates sample quality. Hence, in practice we choose $T$ away from extremal values  and hence too small $\beta$.

    Empirically, the proposed alternating procedure converged in a small number of outer iterations (five in most experiments), consistently yielding stable solutions. We note that the absence of a formal convergence proof concerns the outer alternating loop only: the inner SB subproblem is solved via LightSB-M \citep{pmlr-v235-gushchin24a}, which carries theoretical guarantees.

\subsection{Inference}

Once the drift $s_{\theta}^K$ and the transport map $\Zc_{\tilde{\theta}}^K$ are trained, one can generate new samples from $\mu_T$ by first computing $Y_0 = \Zc_{\tilde{\theta}}^K(X_0) \sim \msY_0 \# \mu_0$, where $X_0 \sim \mu_0$ is an out-of-sample point. Then, sample $Y_T \sim \msY_T \# \mu_T$ according to the learned coupling $\pi_{v_{\theta}}(Y_T \,|\, Y_0)$ given by~\eqref{eq:plan_sb}, and recover $X_T = Y_T + \frac{1}{\beta} \, s_{\theta}^K(T, Y_T) \sim \mu_T$. Alternatively, one could simulate the SDE $\d Y_t = s_{\theta}^K(t, Y_t)\, \d t + \sqrt{\varepsilon}\, \d W_t$, using a numerical SDE solver (e.g., the Euler--Maruyama scheme), but this approach is generally more time-consuming and introduces additional discretization errors.

Note that the drift $s_{\theta}$ defined in~\eqref{eq:drift_lightsbb} is not well-defined at $t = T$. Nevertheless, by continuity of $\phi = \log h$ with respect to time, we can instead approximate $X_{\tilde{T}} = Y_{\tilde{T}} + \frac{1}{\beta} \, s_{\theta}^K(\tilde{T}, Y_{\tilde{T}})$, where $\tilde{T} = T - \delta$ for some small $\delta > 0$.

\section{Numerical Experiments}

In this section, we present numerical experiments to evaluate the proposed algorithm on both univariate and multivariate datasets. We also provide a comparative analysis against state-of-the-art (SOTA) generative models. In all our experiments, we have used $T=1$.

\subsection{Illustrative Examples}

\begin{figure*}[!t]
    \centering
    
    \begin{subfigure}[t]{0.49\textwidth}
        \centering
        \raisebox{2mm}{
        \includegraphics[width=\linewidth]{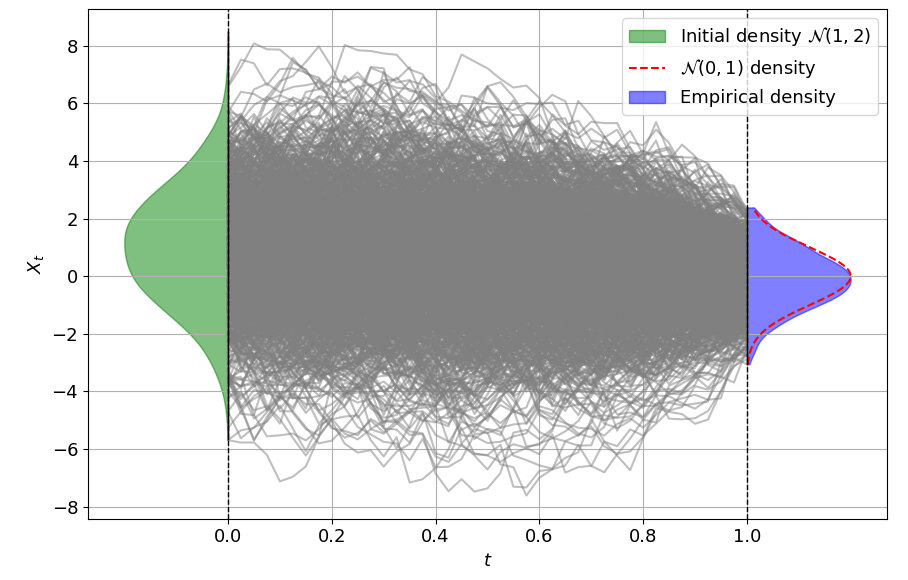}
        }
        \caption{}
    \end{subfigure}
    \hfill
    \begin{subfigure}[t]{0.49\textwidth}
        \centering
        \raisebox{0mm}{             \includegraphics[width=\linewidth]{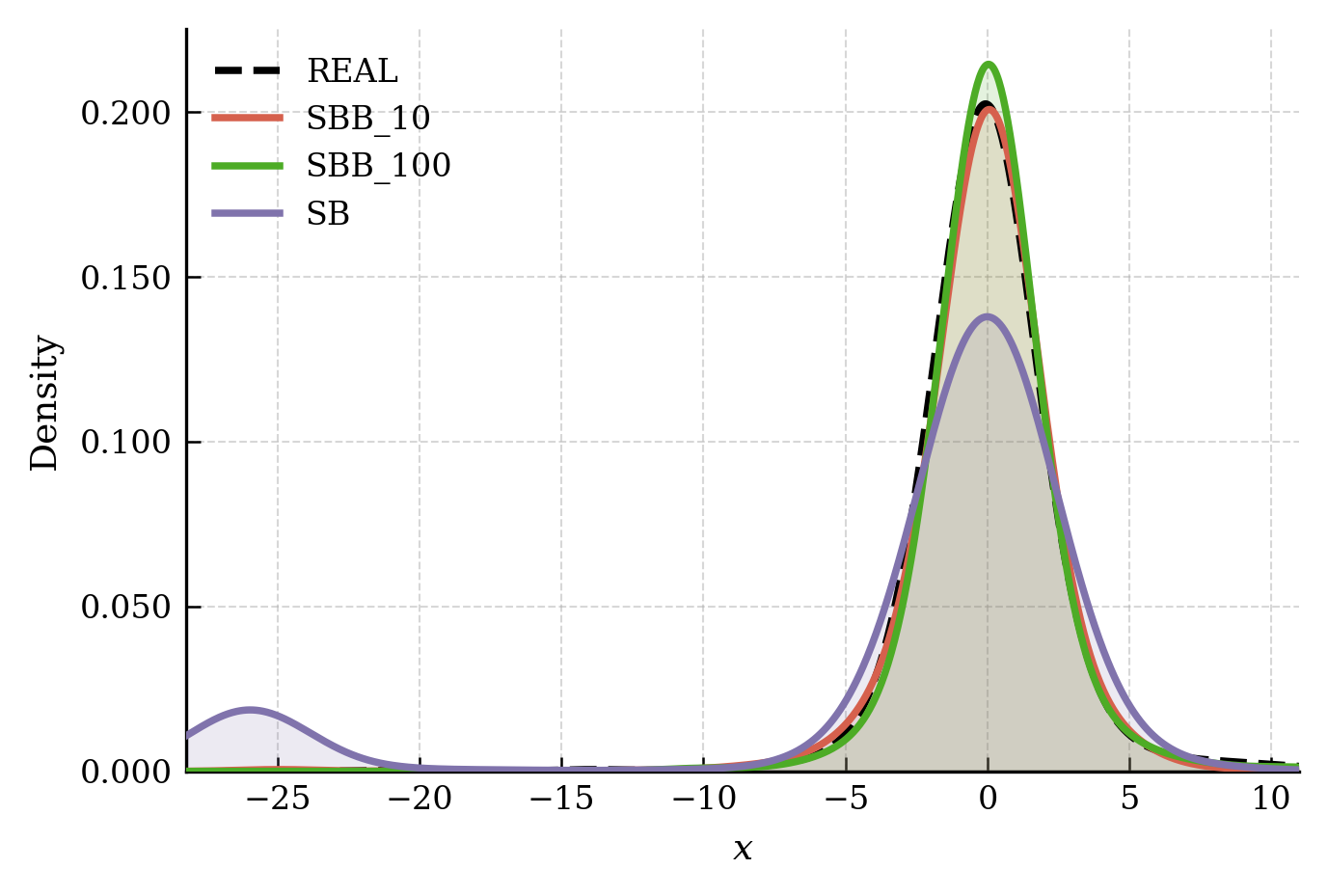}
        }
        \caption{}
    \end{subfigure}
    
    \caption{Transport interpolation using SBB: $\mathcal{N}(1,2) \xrightarrow{} \mathcal{N}(0,1)$ (\textit{left}) and $\mathcal{N}(0,1) \xrightarrow{} \mathcal{T}(2)$ (\textit{right}).}
    \label{fig:transport_toy}
    
    \vspace{-4mm}
\end{figure*}

We propose to use the SBB framework to transport between two simple distributions.
First, we apply SBB between $\mu_0 = \mathcal{N}(1,2)$ and $\mu_T = \mathcal{N}(0,1)$ with parameters $\beta = 10$, $K = 5$, and $M_{\text{samples}} = 2000$.
Figure~\ref{fig:transport_toy}(a) displays the trajectories generated by SBB for this case.

We then consider a more challenging setting involving heavy-tailed distributions.
One of the main advantages of SBB over the classical SB is that it removes the requirement $\mathrm{KL}(\mathbb{P}|\mathbb{W}^\varepsilon) < \infty$. To illustrate this property, we consider $\mu_0 = \mathcal{N}(0,1)$ and $\mu_T = \mathcal{T}(2)$. When the reference marginal at time $T$ is Gaussian, the KL divergence between $\mu_T = \mathcal{T}(2)$ and the Wiener measure diverges, since the integrand behaves as $x^2 p_{\text{Student}}(x) \sim 1/|x|$ for large $|x|$, leading to a logarithmic divergence. Consequently, no finite-entropy SB exists between these marginals under a Brownian prior.

Figure~\ref{fig:transport_toy}(b) illustrates the practical implications of this issue. Numerical approximations of the classical SB produce clear artifacts, including a spurious mode around $x \approx -25$ that is absent from the true distribution, as well as a significant underestimation of the peak density. In contrast, LightSBB-M with $\beta = 10$ and $\beta = 100$ accurately recovers the target $\mathcal{T}(2)$. This demonstrates that jointly controlling drift and volatility enables SBB to handle heavy-tailed distributions for which the classical SB problem is ill-posed.

\subsection{Quantitative Evaluation on Low-Dimensional Datasets}

We provide a quantitative evaluation of the proposed SBB method on low-dimensional datasets, namely \textbf{8gaussians} and \textbf{moons} ($d=2$), and compare it against several SOTA baselines.  
These include alternative SB solvers \cite{pmlr-v235-gushchin24a, NEURIPS2023_c428adf7, tong2024simulationfree, de2021diffusion} and flow-based generative approaches (ODE) \cite{tong2024improving, lipman2023flow, liu2022rectified}.  
Each model transports $10{,}000$ samples from the source to the target distribution, and performance is assessed using the 2-Wasserstein distance.

The 2-Wasserstein distance between two probability measures $q_0$ and $q_1$ on a metric space $(\mathcal{X}, d)$ is defined as
\[
\mathcal{W}_2(q_0, q_1)
  = \left(
    \inf_{\gamma \in \Gamma(q_0, q_1)}
    \int_{\mathcal{X} \times \mathcal{X}}
      \|x - y\|^2 \, d\gamma(x,y)
    \right)^{1/2},
\]
where $\Gamma(q_0, q_1)$ denotes the set of all couplings of $q_0$ and $q_1$.  
In our experiments, we aim to minimize this distance between samples generated by the SBB model and those drawn from the ground-truth distribution.  
All results are averaged over five seeds, and we report both the mean and standard deviation. As shown in Table~\ref{tab:w2_results_clean}, the proposed SBB framework consistently achieves the lowest $\mathcal{W}_2$ distances across all benchmark tasks. It outperforms both SB-based methods (e.g., LightSB-M) and diffusion/flow-based baselines, yielding on average a $\sim19\%$ improvement in transport accuracy, while reducing the variance of the results. These results highlight the efficiency and stability of SBB in modeling complex multimodal and non-Gaussian distributions such as the \textit{moons} $\rightarrow$ 8-\textit{gaussians} task. Note that FM is incompatible with the latter as it requires a Gaussian source distribution. 

We further analyze the influence of $\beta$ and the convergence behavior of the outer loop in Appendix~\ref{sec:quant_setup}, including practical guidelines for selecting $\beta$.

\begin{table}[t!]
\caption{2-Wasserstein distances ($\mathcal{W}_2$) on synthetic datasets (lower is better). The best results are highlighted in \textbf{bold}. *Indicates results taken from \cite{tong2024simulationfree}}
\centering
\resizebox{\columnwidth}{!}{%
\begin{tabular}{lccc}
\toprule
 & \multicolumn{3}{c}{$\mathcal{W}_2$ ($\downarrow$)} \\
\cmidrule(lr){2-4}
\textbf{Algorithm} & $\mathcal{N}\!\rightarrow\!8\text{gaussians}$ & moons $\!\rightarrow\!8\text{gaussians}$ & $\mathcal{N}\!\rightarrow\!\text{moons}$ \\
\midrule
$[\text{SF}]^2$M-Exact* &0.275$\pm$0.058 & 0.726$\pm$0.137 & 0.124$\pm$0.023 \\
$[\text{SF}]^2$M-I*     & 0.393$\pm$0.054 & 1.482$\pm$0.151 & 0.185$\pm$0.028 \\
DSBM-IPF* & 0.315$\pm$0.079 & 0.812$\pm$0.092 & 0.140$\pm$0.006 \\
DSBM-IMF* & 0.338$\pm$0.091 & 0.838$\pm$0.098 & 0.144$\pm$0.024 \\
DSB*      & 0.411$\pm$0.084 & 0.987$\pm$0.324 & 0.190$\pm$0.049 \\
LightSB-M& 0.339$\pm$0.099 & 0.295$\pm$0.051 & 0.201$\pm$0.042 \\
\textbf{SBB (ours)}     &  \textbf{0.241$\pm$0.083} & \textbf{0.201$\pm$0.034} & \textbf{0.109$\pm$0.014} \\
\midrule
OT-CFM*   & 0.303$\pm$0.043 & 0.601$\pm$0.027 & 0.130$\pm$0.016 \\
SB-CFM*   & 2.314$\pm$2.112 & 0.843$\pm$0.079 & 0.434$\pm$0.594 \\
RF*       & 0.421$\pm$0.071 & 1.525$\pm$0.330 & 0.283$\pm$0.045 \\
I-CFM*    & 0.373$\pm$0.103 & 1.557$\pm$0.407 & 0.178$\pm$0.014 \\
FM*       & 0.343$\pm$0.058 & --- & 0.209$\pm$0.055 \\
\bottomrule
\end{tabular}
}
\label{tab:w2_results_clean}
\end{table}

\section{Evaluation on Unpaired Image-to-Image Translation}

To evaluate the generative capabilities of our model we consider the task of
unpaired image‑to‑image translation \cite{zhu2017unpaired} on subsets of the FFHQ dataset at a resolution of $1024\times1024$ pixels \cite{karras2019style}.  The source distribution $p_{0}$ comprises adult faces, while the target distribution $p_{T}$ consists of child faces.  Since the two domains are not paired, we adopt a cycle‑consistent framework in the latent space of an Adversarial Latent AutoEncoder (ALAE) \cite{pidhorskyi2020adversarial}.  Each image $\mathbf{x}\in
\mathbb{R}^{3\times1024\times1024}$ is first encoded by the \textit{pretrained} ALAE encoder $E:\mathbb{R}^{3\times1024\times1024}\rightarrow\mathbb{R}^{512}$, producing a
latent code $\mathbf{z}=E(\mathbf{x})\in\mathbb{R}^{512}$. Then, we train our SBB framework in that latent space, and decode the final latent output using the \textit{pretrained} ALAE decoder $D:\mathbb{R}^{512}\rightarrow \mathbb{R}^{3\times1024\times1024}$, yielding the translated images. This setup allows us to assess the fidelity of the generated child faces and the preservation of
identity-related attributes without requiring paired supervision.

\begin{figure*}[h]
	\centering
	\begin{minipage}[b]{.49\textwidth}
		\centering
		\includegraphics[width=\textwidth]{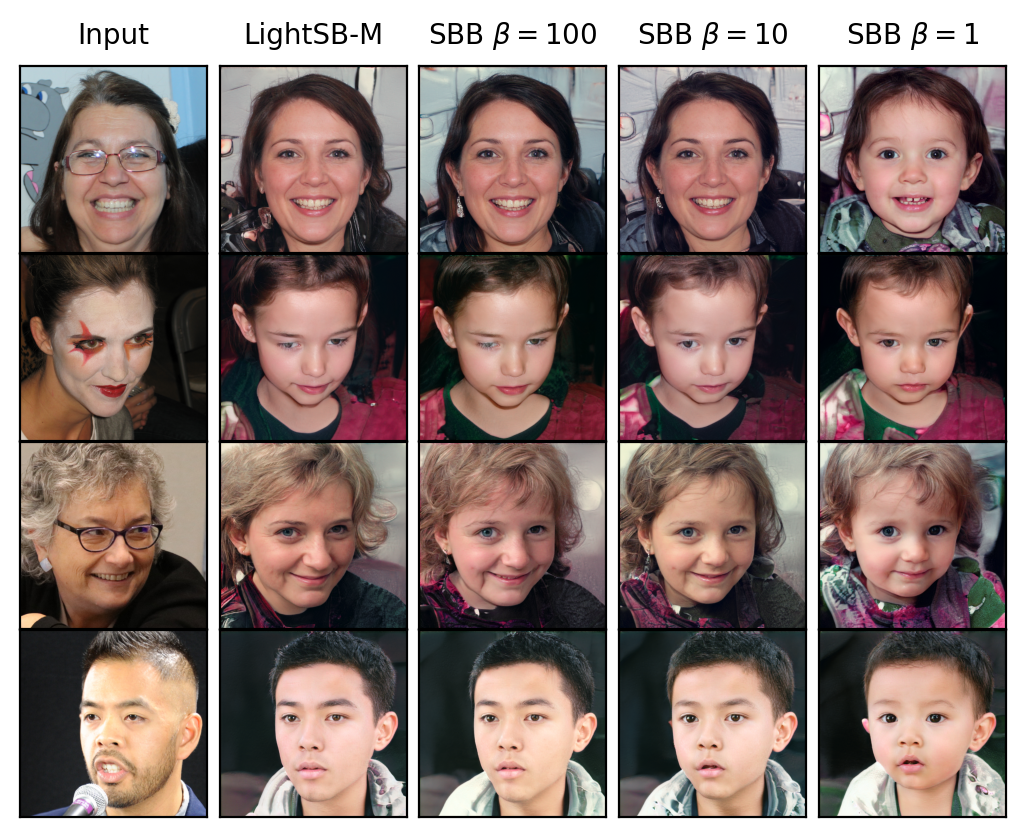}
		\subcaption{$\eps=0.1$}
	\end{minipage}
	\hfill
	\begin{minipage}[b]{.49\textwidth}
		\centering
		\includegraphics[width=\textwidth]{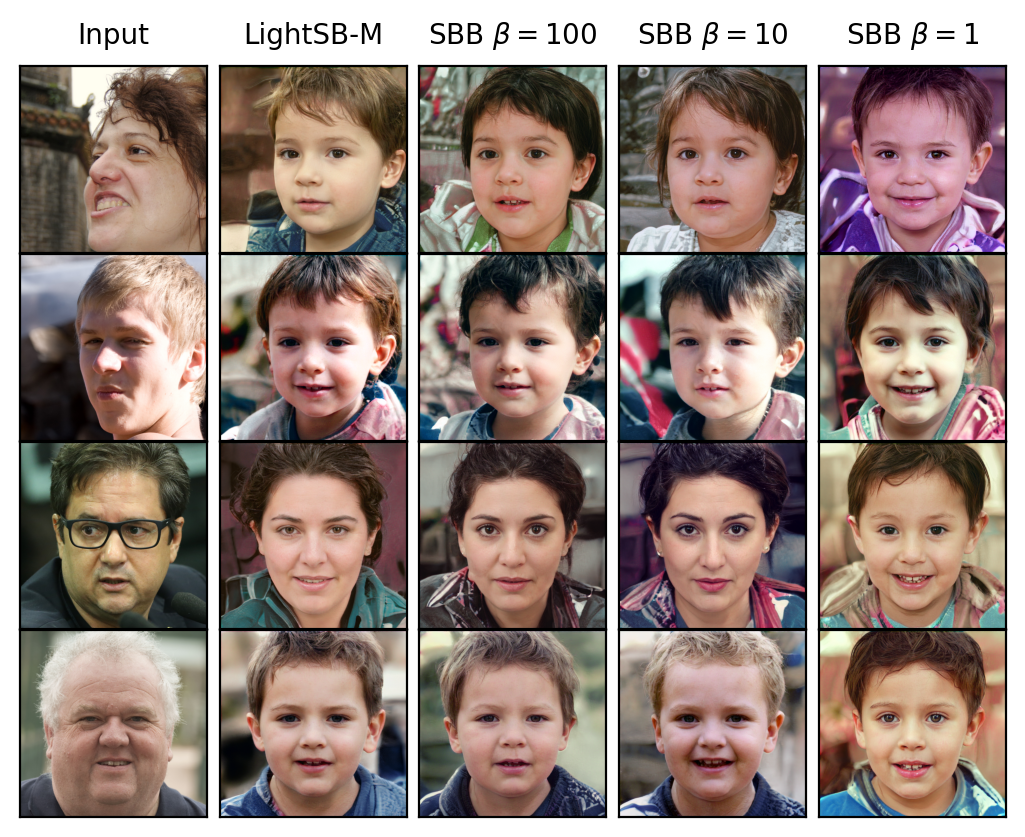}
		\subcaption{$\eps=1$}
	\end{minipage}
	\caption{Comparison between our framework SBB and the benchmark LightSB-M. The left column shows the input image.}
	\label{fig:alae_plots}
\end{figure*}

Figure~\ref{fig:alae_plots} compares our SBB framework with the usual SB approach using the LightSB‑M baseline across several $(\beta,\eps)$ settings, as LightSB-M is the current state-of-the-art on this task and already provides an extensive comparison against further baselines such as DSBM~\cite{NEURIPS2023_c428adf7} in their original  \cite{pmlr-v235-gushchin24a}. 
For the low‑noise regime $\eps=0.1$, SBB with small $\beta$ (e.g., $\b=1,10$ attains higher visual quality and better fidelity than LightSB‑M, as some images are clearly not of children. Moreover, when the noise level is increased to $\eps=1$,  where classical SB exhibits high variance, we observe that small $\b$ values lead to greater diversity in the generated outputs. Finally, we also demonstrate that for large values of $\b$ (i.e., $\b = 100$), the behavior closely matches that of the standard SB, in agreement with the theoretical predictions.

\paragraph{Quantitative evaluation.}
To complement the qualitative comparison, we evaluate age translation quality and identity preservation quantitatively. Age is estimated using InsightFace pre-trained model, and identity similarity is measured by cosine similarity between input and output latent codes. Results are reported in Table~\ref{tab:alae_quant} for both LightSBB-M and LightSB-M across $\varepsilon \in \{0.1, 1\}$.

\begin{table}[t]
\centering
\caption{Quantitative evaluation of adult$\,\to\,$child translation. Age estimated by InsightFace; identity similarity by ArcFace cosine similarity. $\uparrow$/$\downarrow$ indicate higher/lower is better.}
\label{tab:alae_quant}

\resizebox{\columnwidth}{!}{%
\begin{tabular}{lcccc}
\toprule
& \multicolumn{2}{c}{$\varepsilon = 0.1$} & \multicolumn{2}{c}{$\varepsilon = 1$} \\
\cmidrule(lr){2-3}\cmidrule(lr){4-5}
Metric & LightSBB-M & LightSB-M & LightSBB-M & LightSB-M \\
\midrule
Avg.\ age $\downarrow$        & $\mathbf{20.3 \pm 12.3}$ & $27.1 \pm 13.7$ & $\mathbf{12.5 \pm 8.3}$ & $27.2 \pm 13.0$ \\
Age $\leq 18$ (\%) $\uparrow$ & $\mathbf{41.2}$          & $20.7$          & $\mathbf{76.0}$         & $19.3$          \\
Identity sim.\ $\uparrow$     & $0.83 \pm 0.05$          & $\mathbf{0.88 \pm 0.05}$ & $0.81 \pm 0.05$ & $\mathbf{0.85 \pm 0.05}$ \\
\bottomrule
\end{tabular}
}
\end{table}

LightSBB-M substantially improves age translation quality: the fraction of outputs estimated as under $18$ years old is roughly doubled at $\varepsilon=0.1$ (41.2\% vs.\ 20.7\%) and nearly quadrupled at $\varepsilon=1$ (76.0\% vs.\ 19.3\%), with a consistently lower average estimated age across both settings. LightSBB-M shows a modest decrease in identity similarity relative to LightSB-M (approximately 0.04--0.05 cosine similarity units). This is the expected trade-off: LightSB-M retains more adult facial structure and therefore stays closer to the input embedding, while LightSBB-M produces genuinely more child-like faces whose geometry departs further from the adult input. A method that perfectly preserves identity would by definition fail at the translation task.

We also evaluate perceptual
quality using the \textit{Fr\'echet Inception Distance} (FID), computed on decoded
pixel-space images using Inception-v3 features. At $ \beta=1, \varepsilon = 0.1$,
LightSBB-M achieves a FID of $20.9$, comparable to the $21.6$ obtained by
LightSB-M. This confirms that the improved age translation provided by
our framework does not come at the cost of image fidelity: the additional
volatility guides the transport toward genuinely younger faces while preserving perceptual quality.

\begin{figure}[H]
	\centering
	\begin{minipage}[b]{0.23\textwidth}
		\centering
		\includegraphics[width=1.05\textwidth]{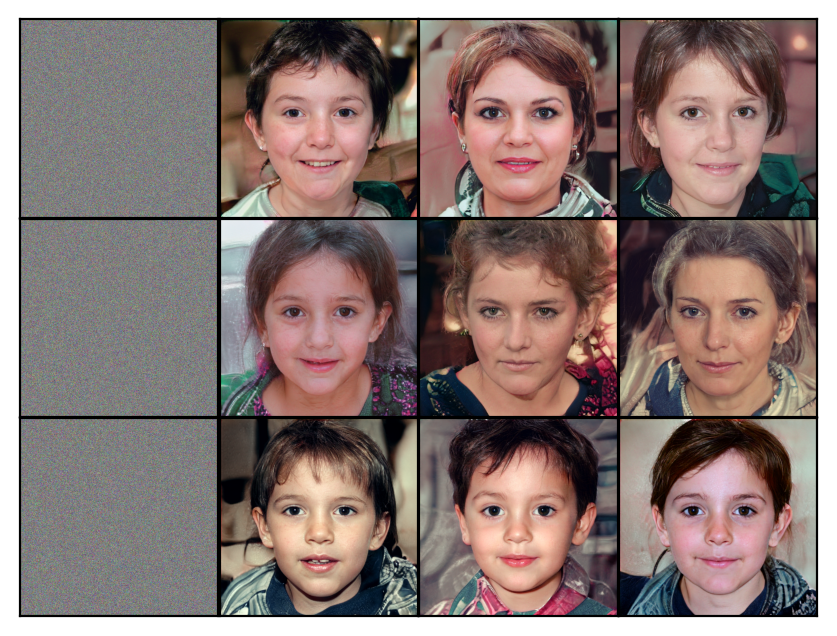}
		\subcaption*{$Y_T$}
	\end{minipage}
	\hfill
	\begin{minipage}[b]{0.23\textwidth}
		\centering
		\includegraphics[width=1.05\textwidth]{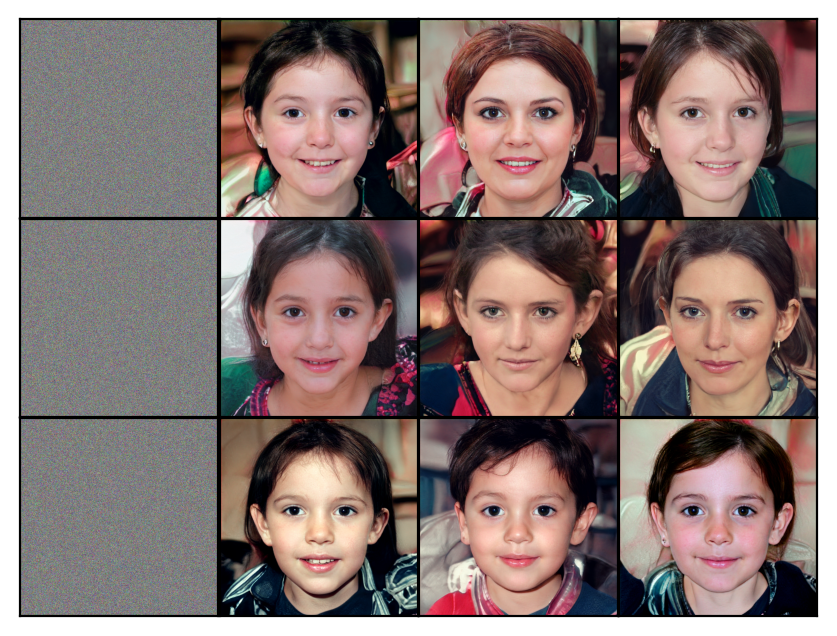}
		\subcaption*{$X_T$}
	\end{minipage}
	\caption{Comparison between $Y_T$ and $X_T$ (\textit{right}) with $\beta=5$. 
		The left column shows the input image, and three representative samples are reported for each method.}
	\label{fig:noise_child}
\end{figure}

We then propose the experiment illustrated in Figure~\ref{fig:noise_child}, which generates a child image directly from noise. Unlike conventional diffusion models that rely on a backward–forward sampling \cite{song2019generative, song2021scorebased} scheme, our approach proceeds in a single forward pass, thereby avoiding the costly reverse diffusion step and error accumulation.
We also present the comparison between the $Y$ and $X$ process, where we observe that the inverse sample $X_T$ significantly improves sample quality and corrects potential errors of $Y_T$. This improvement is evident both in terms of visual fidelity and distributional alignment, as $Y_T$ does not always give child images.

\begin{figure}[H]
	\centering
	\includegraphics[width=1\columnwidth]{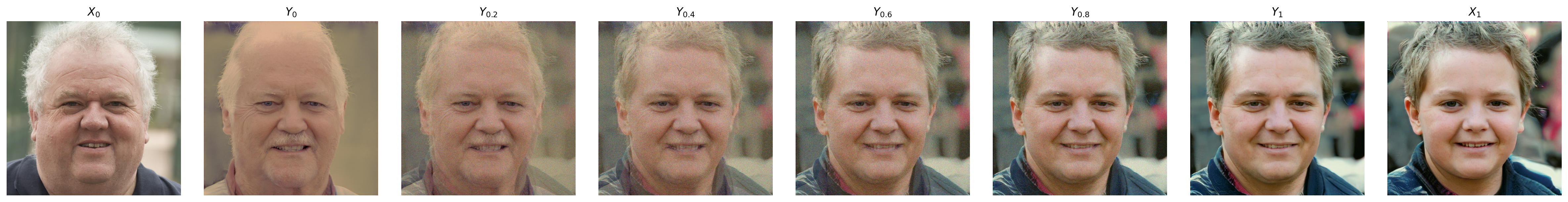}
	\caption{Trajectory from $X_0 \sim p_0$ to $X_1 \sim p_1$ with the underlying $Y$ process.}
	\label{fig:full_framework}
\end{figure}

Finally, Figure~\ref{fig:full_framework} depicts the complete pipeline from the source distribution $X_0 \sim p_0$ to the target distribution $X_1 \sim p_1$ together with the underlying $Y$‑process defined by our framework. The intermediate states $Y_t$ are obtained via bridge matching Equation~\eqref{eq:brownian_bridge} between the endpoints $Y_0$ and $Y_T$. Note that this bridging occurs on the $Y$ process, not on $X$, since $Y$ is a SB

\section{Conclusion}

\textbf{Potential impact.}  
Our primary contribution is an algorithm that solves the SBB problem for generative modeling.  Building on existing methods for the classical SB, we derive an efficient procedure to compute the optimal transport plan $\P^{SBB}$.  The framework includes a tunable parameter $\beta$ that balances drift and volatility; we empirically demonstrate how $\beta$ impacts the sample diversity and fidelity.  By incorporating stochastic volatility, the method can accommodate a broader class of target distributions—including those with heavy tails—beyond the restrictive assumptions of the standard SB.  Consequently, the generated synthetic data exhibit greater variability while maintaining high fidelity as governed by $\beta$.

\textbf{Limitations and future work.}  
Computing $\P^{SBB}$ requires iterating over the transport map, which can be computationally demanding.  In our experiments we limited the iteration count to $K=5$ as it seems to converge, yet the algorithm’s convergence has not been formally established.  Future research should (i) develop tighter iteration‑complexity bounds, (ii) provide a rigorous convergence proof, and (iii) explore acceleration techniques (e.g., stochastic approximations or multigrid schemes) to reduce the number of required iterations.  
A natural extension of the method is to apply the SBB framework to time‑series data.  While the SB problem has already been employed for sequential data \cite{hamdouche2023generativemodelingtimeseries, Alouadi_2025}, the stochastic‑volatility extension proposed here has not.  Incorporating the controlled volatility could markedly improve the realism of generated time series, particularly in financial domains where heteroskedasticity is prevalent.

 \bibliography{example_paper}
 \bibliographystyle{icml2026}

\newpage
\appendix
\onecolumn

\section{Proof of Proposition~\ref{prop:envelope}}
\label{app:proof_envelope}

\textit{1.} Follows directly from \citep[Remark~2.3]{henrylabordere2026bridgingschrodingerbasssemimartingale}: for every $\phi \in C_w$, $\tilde\phi := \Tc_\beta^- \circ \Tc_\beta^+[\phi] \in C_w^{\mathrm{conv}} = \mathcal{C}_w \cap \mathcal{C}^{\text{conv}}, \quad \text{with } \mathcal{C}^{\text{conv}} := \{ \phi \in C^0(\mathbb{R}^d) : \phi \text{ is } \beta\text{-convex} \}$ and $J(\tilde\phi) \ge J(\phi)$.

\textit{2.} Let $v_t^{\tilde\phi}(x) := \Tc_\beta^+[u_{T-t}^{\tilde\phi}](x)$. Since $\tilde\phi \in C_w^{\mathrm{conv}}$, \citep[Lemma~5.1]{henrylabordere2026bridgingschrodingerbasssemimartingale} yields $v^{\tilde\phi} \in C^{1,2}$ and
\[
D^2 v_t^{\tilde\phi}(x) - \beta I_d = -\beta^2 \Big(\beta I_d + D^2 u_{T-t}^{\tilde\phi}(\msY_t^{\tilde\phi}(x))\Big)^{-1}.
\]
The same lemma gives the lower Hessian bound $D^2 u_{T-t}^{\tilde\phi}(y) + \kappa(t) I_d \succeq 0$ with $\kappa(t) = \frac{\beta}{1+\beta(T-t)}$, so $\beta I_d + D^2 u_{T-t}^{\tilde\phi}(y) \succeq (\beta - \kappa(t)) I_d \succ 0$ for $t < T$, giving $D^2 v_t^{\tilde\phi}(x) < \beta I_d$.

\textit{3.} Follows from \citep[Theorem~3.2(c)]{henrylabordere2026bridgingschrodingerbasssemimartingale} under $\beta T > 1$. The positivity $D\msY_t^{\tilde\phi}(x) > 0$ follows from $D\msY_t^{\tilde\phi}(x) = \beta(\beta + D^2 u_{T-t}^{\tilde\phi}(\msY_t^{\tilde\phi}(x)))^{-1} \succ 0$.

The volatility bound with parameter $\beta - \delta$ follows from $I_d - \frac{1}{\beta} D^2 v_t \succeq \frac{\delta}{\beta} I_d$ and the exact control formula $\sigma^*(t,x) = \sqrt{\varepsilon}(I_d - \frac{1}{\beta} D^2 v_t(x))^{-1}$. \qed

\section{Exploring Alternative Algorithms}
\label{sec:other_algo}

In this section, we are exploring other potential algorithms to solve the SBB problem.

\subsection{LightSBB-M for $\b$ Large}

We recall that the objective of Algorithm~\ref{alg:lightsbb} is to estimate both the drift of the process $Y$ and the transport map $\msY_t(X_t) = X_t - \frac{1}{\beta} \eps \nabla_y \log h_t^*(\msY_t(X_t))$ for $t \in [0, T]$. One may notice that for large $\beta$, we can make the following approximation of the transport map:
\beq
\label{eq:map_beta_large}
\msY_t(X_t) = X_t - \frac{1}{\beta} \eps \nabla_y \log h_t^*(\msY_t(X_t)) \simeq X_t - \frac{1}{\beta} \eps \nabla_y \log h_t^*(X_t)
\enq

Approximation \ref{eq:map_beta_large} hence provides an explicit expression, given the drift, of the transport map and avoid computing the inverse of $\msX$. One can simply sample from $\msY_t \# \mu_t$, at iteration $k$, using 
\beq
\label{eq:map_beta_large_sample}
\begin{cases}
	\msY_0^k(x_0) = y_0 = x_0 - \frac{1}{\beta} s_{\t}^k(0, x_0), \\
	\msY_T^k(x_T) = y_T = x_T - \frac{1}{\beta} s_{\t}^k(T, x_T)
\end{cases}
\enq

Then, Algorithm~\ref{alg:lightsbb} simply becomes as described in Algorithm~\ref{alg:sbb_algo_beta_large}.

\begin{algorithm}[H]
	\caption{LightSBB-M for $\beta$ large}\label{alg:sbb_algo_beta_large}
	\begin{algorithmic}
		\STATE {\bfseries Input:} Samples $(x_0^m, x_T^m)_{m \leq M} \sim(\mu_0,\mu_T)$, $\theta = \{\alpha_j, \mu_j, \Sigma_j \}_{j \leq J}$, $\beta>0$ large, $K >0$

		\STATE \textbf{Initialization:} Start with $\msY^0 = I_d$
		\FOR{$k=0, \cdots, K-1$}	
		\REPEAT
		\STATE Draw sample batch of pairs $(x_0^n, x_T^n)_{n \leq N}$
		\STATE Compute $\msY_0^k(x_0^n) =y_0^n$ and $\msY_T^k(x_T^n)=y_T^n$ using ~\eqref{eq:map_beta_large_sample}
		\STATE Sample batch $(y_t^n)_{n \leq N} \sim \mathbb{W}_{|y_0, y_1}$ using ~\eqref{eq:brownian_bridge}
		\STATE Compute the drift $s_{\theta}^k$ using ~\eqref{eq:drift_lightsbb} and update $\theta^{k}$ by minimizing ~\eqref{eq:loss_lightsb}
		\UNTIL{convergence}
		\STATE $\theta^{k+1} \leftarrow{\theta^k}$
		\ENDFOR
		\STATE \textbf{Return} $\theta^K$ 
	\end{algorithmic}
\end{algorithm}

Once $s_{\theta}^K$ is trained, one can sample similarly as before. First, compute $Y_0 = X_0 - \frac{1}{\beta} s_{\t}^K(0, X_0) \sim \msY_0 \# \mu_0$ where $X_0 \sim \mu_0$ is an out-of-sample point. Then, sample $Y_T \sim \msY_T \# \mu_T$ according to the learned coupling $\pi_{v_{\theta}}(Y_T \,|\, Y_0)$ given by~\eqref{eq:plan_sb}, and recover $X_T = Y_T + \frac{1}{\beta} \, s_{\theta}^K(T, Y_T) \sim \mu_T$. 

This approximation eliminates the need to learn an explicit transport map, which reduces computational cost and mitigates the accumulation of approximation errors during training.

\subsection{Sinkhorn-based Algorithm}

We introduce a Sinkhorn-based algorithm that iteratively updates the potential function $\phi$, setting $\varepsilon = 1$ for simplicity of notation.  
The algorithm proceeds as follows, starting from $\phi^0 = 0$, for all $0 \leq k \leq K$:

\begin{enumerate}
    \item Compute $h_0^k(y) = \mathbb{E}_{\mathcal{N}(0, T)}(e^{\phi^k(y + Z)}) \simeq \frac{1}{N} \sum_{i=1}^{N} e^{\phi^k(y + Z_i)}$ with $(Z_i)_{i \leq N} \sim \mathcal{N}(0, T)$ i.i.d.

    \item Compute $\msY_0^k(x) = \arg\min_{y\in\R^d} \left[ \log h_0^k(y) + \frac{\beta}{2}\|x-y\|^2 \right]$ for $x \sim \mu_0$.  
    This can be achieved either via grid search, or approximately (for large $\beta$) using $\msY_0^k(x) \simeq x - \frac{1}{\beta} \nabla_y \log h_0^k(x)$, where $\nabla_y \log h_0^k(x)$ is obtained by interpolating $\log h_0^k$ and estimating its gradient via finite differences for all $x \sim \mu_0$.

    \item Compute $\msY_T^k(x) = \arg\min_{y\in\R^d} \left[ \phi^k(y) + \frac{\beta}{2}\|x-y\|^2 \right]$ for $x \sim \mu_T$.  
    Similarly, this can be done either through grid search or, when $\beta$ is large, by approximating $\msY_T^k(x) \simeq x - \frac{1}{\beta} \nabla_y \phi^k(x)$, where the gradient is computed from an interpolated version of $\phi^k$ using finite differences.

    \item Compute $h_T^{k+1}(y) = e^{\phi^{k+1}(y)} = \frac{\mathbb{E}_{\mu_T}\left(\delta_{\msY_T^k(X_T)}(y)\right)}{\mathbb{E}_{\mu_0}\left( \frac{\mathcal{N}_{\msY_0^k(X_0), T}(y)}{h_0^k(\msY_0^k(X_0))} \right)}$ via Monte Carlo estimation using samples from $\mu_T$ and $\mu_0$.

    \item Update $\phi^{k+1} = \log h_T^{k+1}$.
\end{enumerate}

After convergence, $\phi^\star \simeq \phi^{K}$, one can simulate the diffusion  
$\d Y_t = \nabla_y \log h_t^K(Y_t) \d t + \d W_t$  
starting from $Y_0 \sim \msY_0^K \# \mu_0$ obtained in step 2, with $h_t^K = h_T^K * \mathcal{N}_{T-t}$ computed on a grid and interpolated to estimate its score via finite differences. Note that this approach is computationally demanding and does not scale well to high-dimensional settings due to: (i) the need for interpolation, (ii) the grid search procedure, and (iii) the estimation of the density of $\msY_T^k\#\mu_T$ in the numerator of step 4, which can be performed using kernel regression methods.

Alternatively, one may directly simulate the process $X$ as follows:

\begin{itemize}
    \item \textbf{Compute $h_t^{K}(\cdot)$} as in step 4, using $\msY_T^K(\cdot)$ and $\msY_0^K(\cdot)$.
    \item \textbf{Compute $\mathscr{Y}_t^K(\cdot)$} as in step 3.
    \item \textbf{Compute the optimal drift and volatility controls.}  
    The expressions of the optimal drift $\a^K$ and volatility $\s^K$ at time $t$ are given by:
    \[
    \begin{cases}
        \a^K(t,x) = \nabla_y \log h_t^K(\mathscr{Y}_t^K(x))  = \frac{\nabla_y {h_t}^K(\mathscr{Y}_t^K(x))}{{h}_t^K(\mathscr{Y}_t^K(x))}, \\
        \s^K(t,x) = I_d + \frac{1}{\beta} \nabla_y^2 \log h_t^K(\mathscr{Y}_t^K(x))
        = I_d + \frac{1}{\beta} \left( \frac{\nabla_y^2 h_t^K(\mathscr{Y}_t^K(x))}{ h_t^K(\mathscr{Y}_t^K(x))} -  \frac{\nabla_y h_t^K(\mathscr{Y}_t^K(x)) \nabla_y h_t^K(\mathscr{Y}_t^K(x))^T}{ h_t^K(\mathscr{Y}_t^K(x))^2} \right)
    \end{cases}
    \]
    where the derivatives are obtained from the interpolated version of $h_t^K(\cdot)$ using finite differences.
\end{itemize}

\begin{figure}[!t]
    \centering
    \begin{minipage}[t]{0.49\textwidth}
        \centering
        \includegraphics[width=\textwidth]{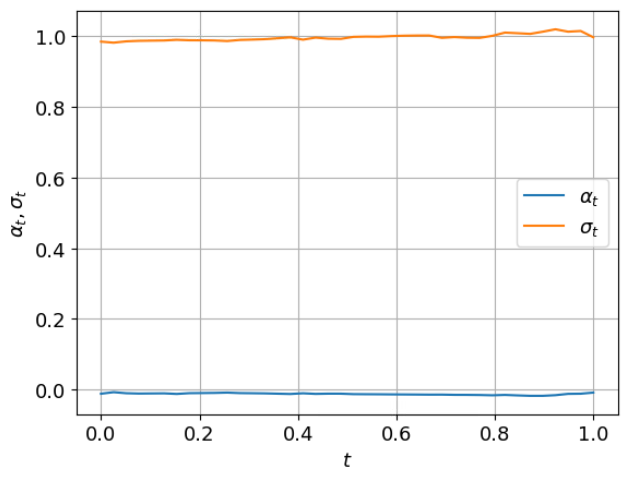}
    \end{minipage}
    \begin{minipage}[t]{0.47\textwidth}
        \centering
        \raisebox{2.3mm}{\includegraphics[width=\textwidth]{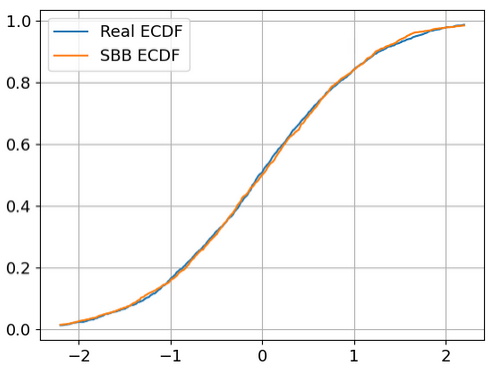}}
    \end{minipage}
    \caption{Estimated optimal drift/volatility over time (\textit{left}) and Empirical CDF comparison between real and SBB-generated samples (\textit{right})}
    \label{fig:ecdf_drift_vol}
\end{figure}

We propose to evaluate this approach by transporting between measures $\mu_0$ and $\mu_T$ for which the optimal drift and volatility are known.  
As an illustrative example, consider $\mu_0 = \delta_{0}$ and $\mu_T = \mathcal{N}(0, T)$.  
In this case, the optimal drift and volatility are given by $\alpha_0^* = 0$ and $\sigma_0^* = I_d$, since $X_T = \alpha_0^* T + \sigma_0^* W_T = 0 + I_d \times \mathcal{N}(0, T) \sim \mathcal{N}(0, T)$.
We perform the experiment using $2000$ real samples drawn from $\mathcal{N}(0, T)$ with parameters $T = 1$, $\beta = 1$, $K = 20$, $N^{\pi} = 1$, $N = 1$, and initialization $\phi^0(y) = \frac{y^2}{4}$.  
The estimated values are $\hat{\alpha}_0 = 0.002$ and $\hat{\sigma}_0 = 1.001$.  
Figure~\ref{fig:ecdf_drift_vol} shows the estimated drift and volatility obtained with SBB across all time steps $t \in [0, T]$, along with the empirical cumulative distribution function (ECDF) computed from both real and generated samples.

\begin{figure*}[!t]
    \centering

    \begin{minipage}[b]{0.48\textwidth}
        \centering
        \includegraphics[width=\textwidth]{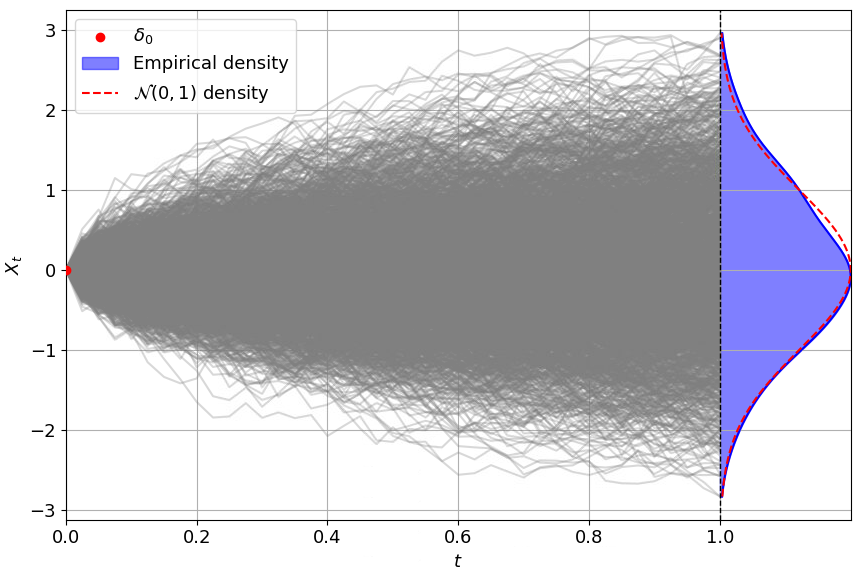}
    \end{minipage}
    \begin{minipage}[b]{0.504\textwidth}
        \centering
        \includegraphics[width=\textwidth]{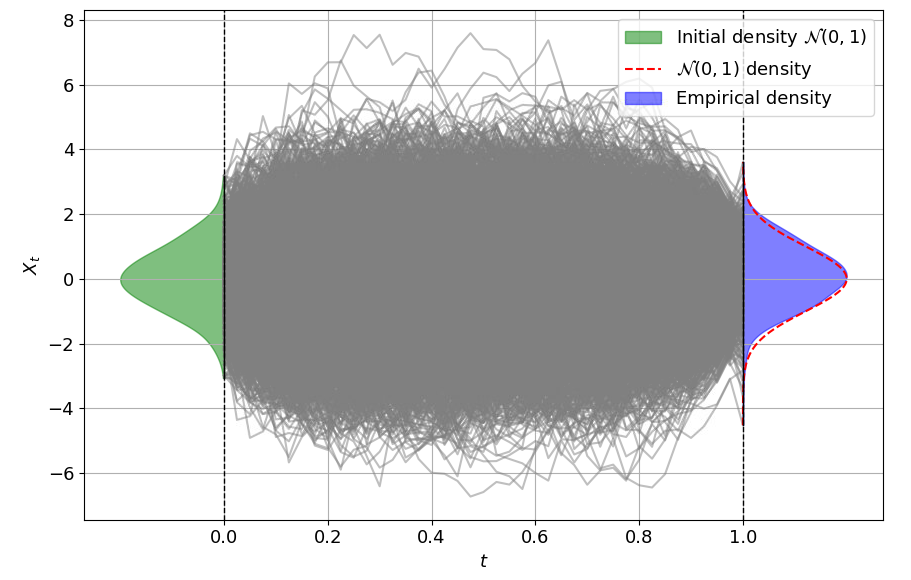}
    \end{minipage}

    \vspace{1em}

    \begin{minipage}[b]{0.482\textwidth}
        \centering
        \includegraphics[width=\textwidth]{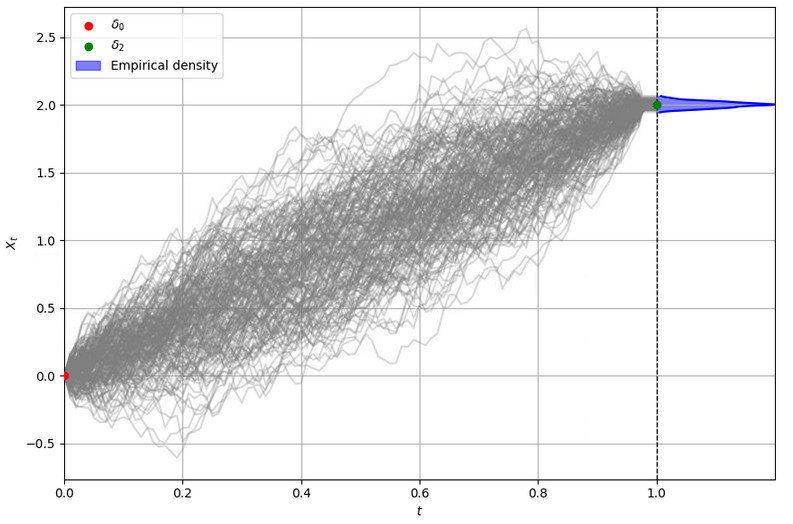}
    \end{minipage}
    \begin{minipage}[b]{0.5\textwidth}
        \centering
        \includegraphics[width=\textwidth]{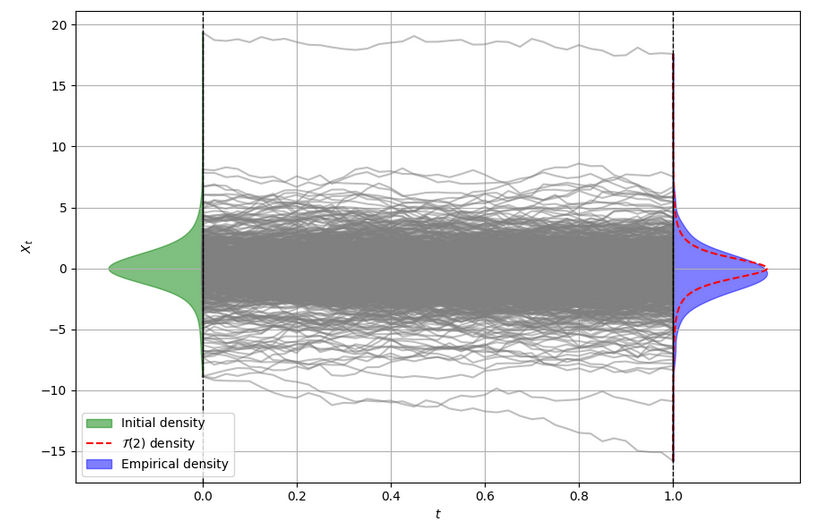}
    \end{minipage}

    \vspace{1em}

    \begin{minipage}[b]{0.5\textwidth}
        \centering
        \includegraphics[width=\textwidth]{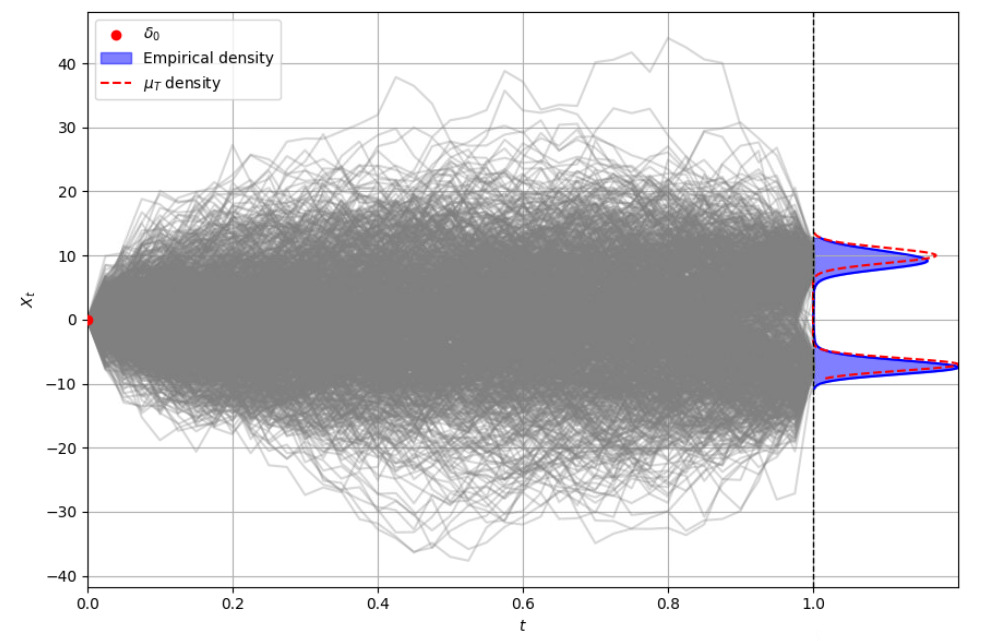}
    \end{minipage}
    \begin{minipage}[b]{0.493\textwidth}
        \centering
        \includegraphics[width=\textwidth]{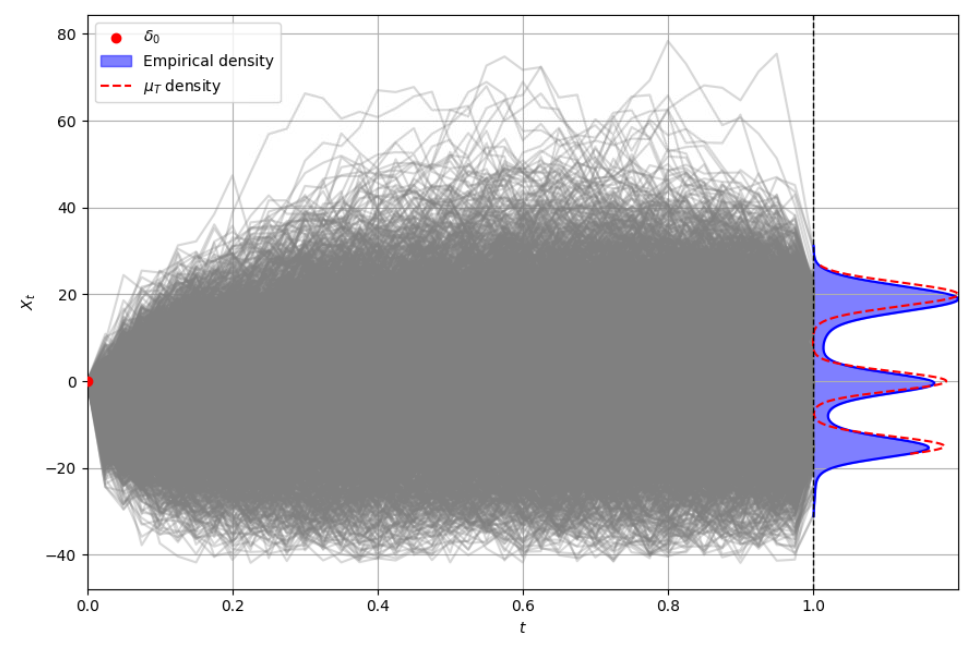}
    \end{minipage}

    \caption{Transport interpolation examples. 
    (\textit{Top}) $\delta_0 \!\rightarrow\! \mathcal{N}(0,1)$ and $\mathcal{N}(0,1) \!\rightarrow\! \mathcal{N}(0,1)$. 
    (\textit{Middle}) $\delta_0 \!\rightarrow\! \delta_2$ and $\mathcal{T}(2) \!\rightarrow\! \mathcal{T}(2)$. 
    (\textit{Bottom}) $\delta_0 \!\rightarrow\!$ multimodal distribution.}
    \label{fig:transport_toy_all}
\end{figure*}

We also present in Figure~\ref{fig:transport_toy_all} illustrative toy examples in one dimension, demonstrating the behavior of the alternative Sinkhorn algorithm for transporting between different probability measures. In these experiments, we set $\beta = 10$, $K = 20$, and $N^\pi = 40$, corresponding to the number of time steps in the Euler scheme. For the density estimation in step 4, we employ the following kernel function:
\beqs
	K_\lambda(x) = \frac{1}{\lambda^{d}} \big(1 - \big\| \tfrac{x}{\lambda} \big\|^2 \big)^2 \, \mathds{1}_{\{\|x\| < \lambda\}}, 
	\quad x \in \R^d,
\enqs
where $\lambda = 0.3$ denotes the kernel bandwidth.

\section{Experimental Setup}

All experiments were conducted using a single NVIDIA A100 SXM4 40 GB GPU. The parameters used throughout this study, unless otherwise stated, are summarized in Table~\ref{tab:parameters_exp}. 

\begin{table}[!ht]
\centering
\caption{Table of parameters and their values}
\label{tab:parameters_exp}
\begin{tabular}{cccccc}
\midrule
$K$ & $T$ & $\tilde{T}$ &  $n_{epoch}$ & Batch Size & $lr$ \\ 
\midrule
$5$ & $1$ & $0.99$ & $15000$ & $512$ & $10^{-3}$ \\
\midrule
\end{tabular}
\end{table}

\subsection{Runtime and Efficiency}
\label{app:runtime}

We now present in Table~\ref{tab:runtime} a comprehensive analysis of both training and inference costs for our algorithm, compared to the LightSB-M baseline. The additional computational cost of LightSBB-M relative to LightSB-M arises from learning the implicit inverse map $\msY_t$, defined as the inverse of $\msX_t(y) = y + \frac{1}{\beta} s_\theta(t,y)$, via the alternating outer loop. In practice this loop requires only $K=5$ outer iterations, and we decrease the number of epochs as $k \leq K$ increases. Wall-clock times are reported on a single NVIDIA A100 GPU; times for LightSBB-M are averaged over all outer iterations $k$.

\begin{table}[h]
\centering
\small
\caption{Wall-clock times on a single NVIDIA A100. ``Inverse/epoch'' refers to training the map $\Zc_{\tilde\theta}$; ``---'' indicates Algorithm~\ref{alg:sbb_algo_beta_large} is used (no inverse network needed). Inference is evaluated on 10{,}000 samples for 2D tasks and 50 samples for the 512-dim latent space.}
\label{tab:runtime}
\begin{tabular}{lcccc}
\toprule
Model & Total & Drift/epoch & Inverse/epoch & Inference \\
\midrule
\multicolumn{5}{l}{\textit{2D $\mathcal{N} \to$ moons}} \\
\midrule
LightSB-M                        & 38s  & 38s & ---  & 8s  \\
LightSBB-M ($\beta=100$)  & 155s & 31s & ---  & 8s  \\
LightSBB-M ($\beta=1$)    & 322s & 31s & 33s  & 8s  \\
\midrule
\multicolumn{5}{l}{\textit{ALAE 512-dim latent space}} \\
\midrule
LightSB-M                        & 38s  & 35s & ---  & 16s \\
LightSBB-M ($\beta=100$)  & 178s & 35s & ---  & 19s \\
LightSBB-M ($\beta=1$)    & 340s & 31s & 53s  & 19s \\
\bottomrule
\end{tabular}
\end{table}

Three observations are in order. First, \textbf{inference cost is essentially identical} across all settings: the additional network $\Zc_{\tilde\theta}$ requires only a single forward pass through a lightweight MLP at inference time. Second, for large $\beta$ (e.g.\ $\beta=100$), the inverse map reduces to the explicit first-order approximation $\msY_t(x) \approx x - \frac{1}{\beta} s_\theta(t,x)$ (Algorithm~\ref{alg:sbb_algo_beta_large}), so no inverse network is needed; the moderate overhead ($\sim$4--5$\times$) comes purely from the $K=5$ outer iterations. Third, for small $\beta$ (e.g.\ $\beta=1$) the inverse map must be learned explicitly, adding the ``Inverse/epoch'' cost. This regime is also where LightSBB-M provides the largest transport quality gains and where classical SB is inapplicable by construction (the finite-entropy constraint $\mathrm{KL}(\mathbb{P}\|\mathbb{W}^\varepsilon) < \infty$ is violated). The training overhead is therefore well-justified by the additional expressivity.

\subsection{Model Architecture}

The transport map $\Zc_{\t}$ is defined by a simple multilayer perceptron (MLP). The network takes as input the time $t \in \R$ and the state $x \in \R^d$. Each input is first processed by a feed-forward network (FFN) comprising a linear layer, a normalization layer, a GELU activation function, and a final linear layer. This projects the inputs into latent spaces of dimension $t_{\text{model}}$ and $d_{\text{model}}$, respectively. The resulting embeddings are then concatenated and passed through a subsequent FFN with a similar structure, which maps the combined representation back to the original space $\R^d$.

\subsection{Quantitative Evaluation and Hyperparameter Analysis}
\label{sec:quant_setup}

	For these experiments, we set the model hyperparameters to $t_{\text{model}}=8$ and
$d_{\text{model}}=32$ with $J=50$ potentials in the Gaussian mixture.  To remain
consistent with the setup in \citep{tong2024simulationfree}, we use $\eps=1$ for all
datasets, except for \textit{moons $\!\rightarrow\!8$ gaussians}, where we set
$\eps=5$.  Table~\ref{tab:all_beta_2d} reports the 2-Wasserstein distances ($\mathcal{W}_2$) obtained for varying values of $\beta$ across all considered datasets.

\begin{table}[H]
\caption{2-Wasserstein distances ($\mathcal{W}_2$) on synthetic datasets (lower is
better) for different $\beta$ values. The best results are highlighted in
\textbf{bold}.}
\centering
\begin{tabular}{lccc}
    \toprule
    & \multicolumn{3}{c}{$\mathcal{W}_2$ ($\downarrow$)} \\
    \cmidrule(lr){2-4}
    \textbf{$\beta$} & $\mathcal{N}\!\rightarrow\!8\text{gaussians}$ & moons
    $\!\rightarrow\!8\text{gaussians}$ & $\mathcal{N}\!\rightarrow\!\text{moons}$ \\
    \midrule
    1    & 4.103$\pm$1.247 & 2.312$\pm$1.089 & 0.927$\pm$0.743 \\
    10   & \textbf{0.241$\pm$0.083} & 0.603$\pm$0.048 & 0.372$\pm$0.032 \\
    50   & 0.277$\pm$0.096 & 0.243$\pm$0.029 & 0.152$\pm$0.024 \\
    100  & 0.330$\pm$0.077 & \textbf{0.201$\pm$0.034} & \textbf{0.109$\pm$0.014} \\
    1000 & 0.462$\pm$0.091 & 0.232$\pm$0.025 & 0.171$\pm$0.037 \\
    $\infty$ & 0.489$\pm$0.084 & 0.334$\pm$0.058 & 0.229$\pm$0.019 \\
    \bottomrule
\end{tabular}
\label{tab:all_beta_2d}
\end{table}

\begin{figure}[ht]
\centering
\begin{subfigure}[t]{0.48\columnwidth}
    \centering
    \includegraphics[width=\linewidth]{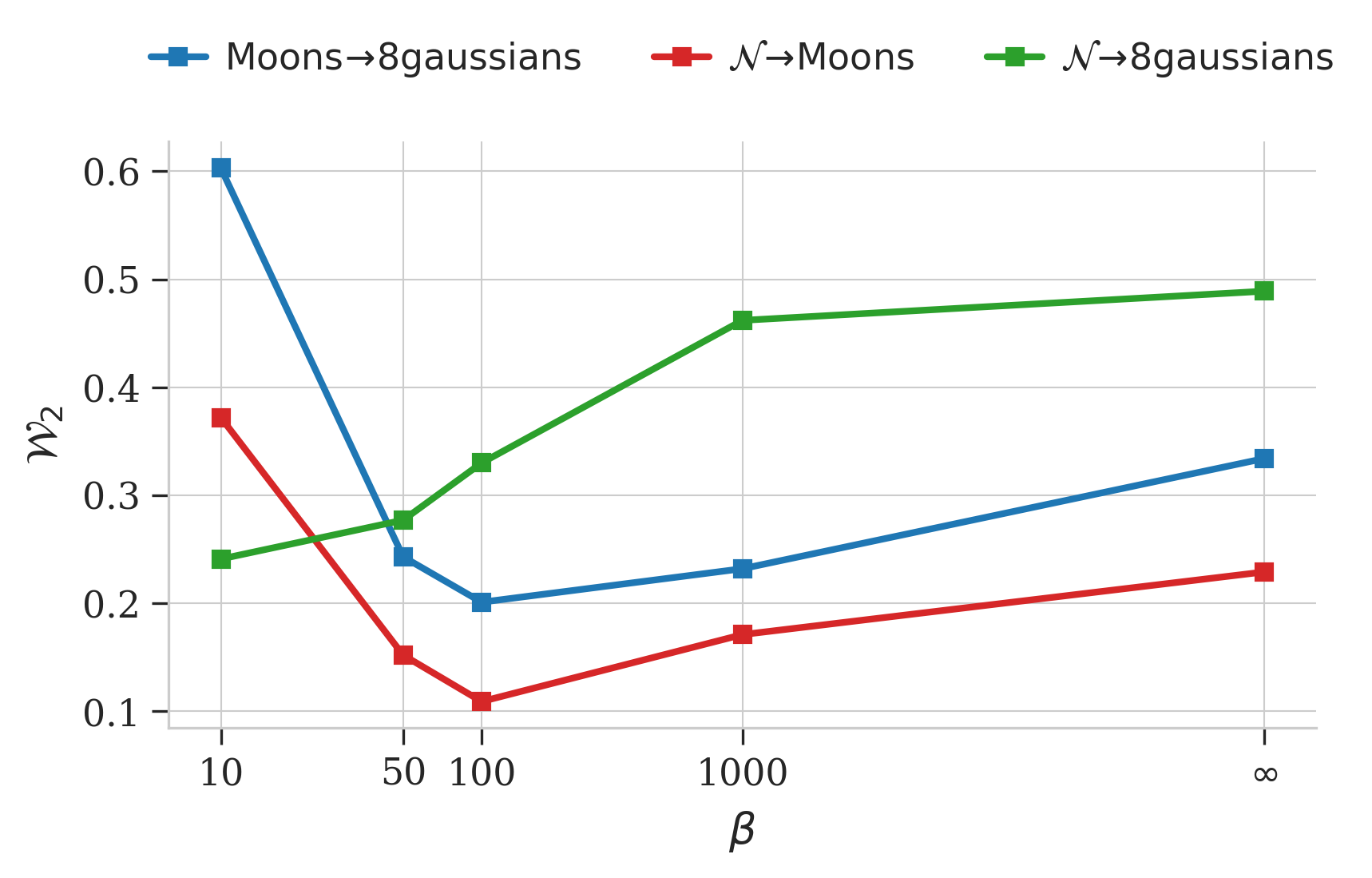}
    \caption{$\mathcal{W}_2$ vs.\ $\beta$}
\end{subfigure}
\hfill
\begin{subfigure}[t]{0.48\columnwidth}
    \centering
    \includegraphics[width=\linewidth]{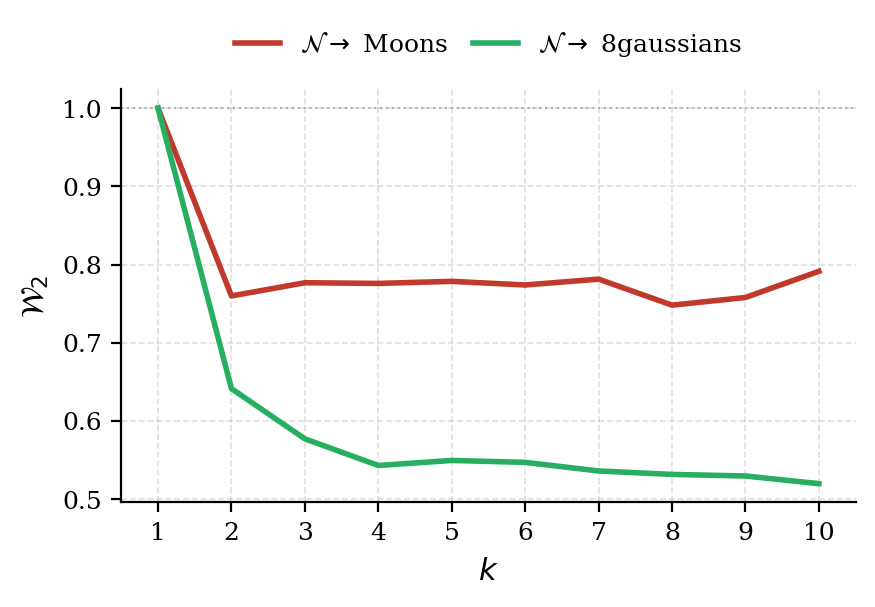}
    \caption{$\mathcal{W}_2$ vs.\ $K$ (normalized)}
\end{subfigure}
\caption{(a) Evolution of the 2-Wasserstein distance for different values of
$\beta$. (b) Normalized 2-Wasserstein distance
$\mathcal{W}_2(k)/\mathcal{W}_2(1)$ as a function of outer iterations $K$ with
$\beta = 100$, showing convergence within 3--5 iterations.}
\label{fig:w2_beta}
\end{figure}

\begin{figure}[t!]
    \centering
    \begin{subfigure}[t]{\textwidth}
        \centering
        \includegraphics[width=\textwidth]{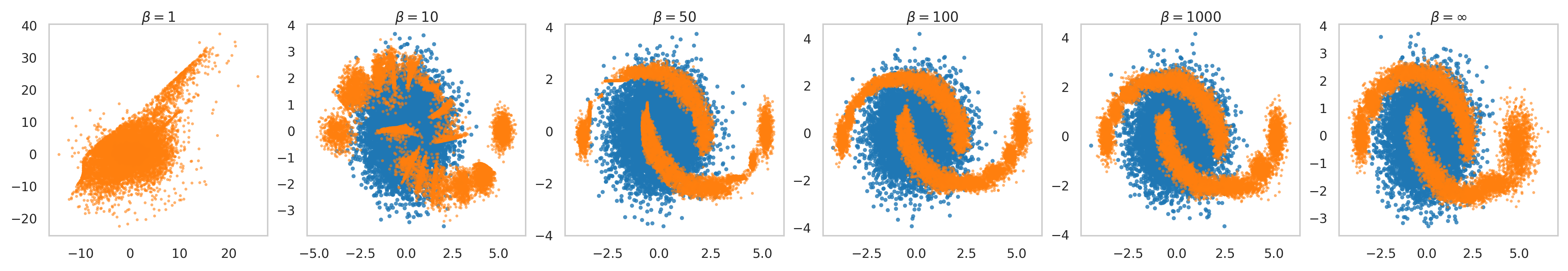}
        \caption{$\mathcal{N}\!\rightarrow\!\text{moons}$}
        \label{fig:nm}
    \end{subfigure}
    \vspace{0.5em}
    \begin{subfigure}[t]{\textwidth}
        \centering
        \includegraphics[width=\textwidth]{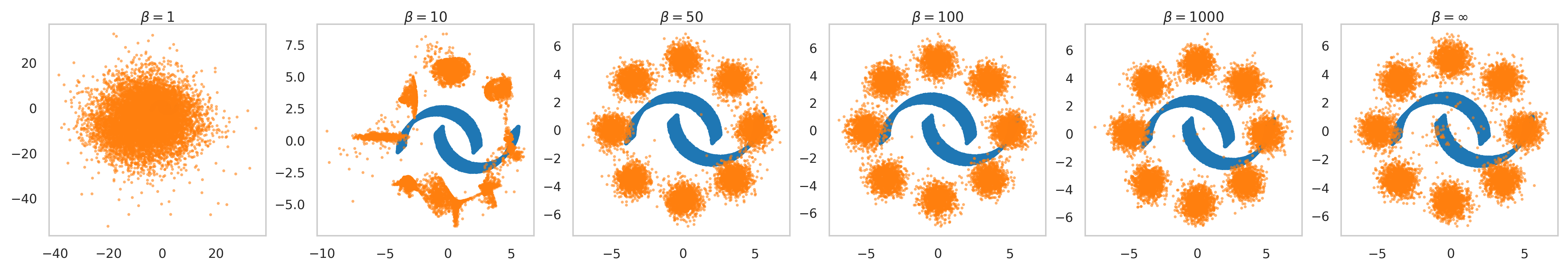}
        \caption{moons $\!\rightarrow\!8\text{gaussians}$}
        \label{fig:m8g}
    \end{subfigure}
    \vspace{0.5em}
    \begin{subfigure}[t]{\textwidth}
        \centering
        \includegraphics[width=\textwidth]{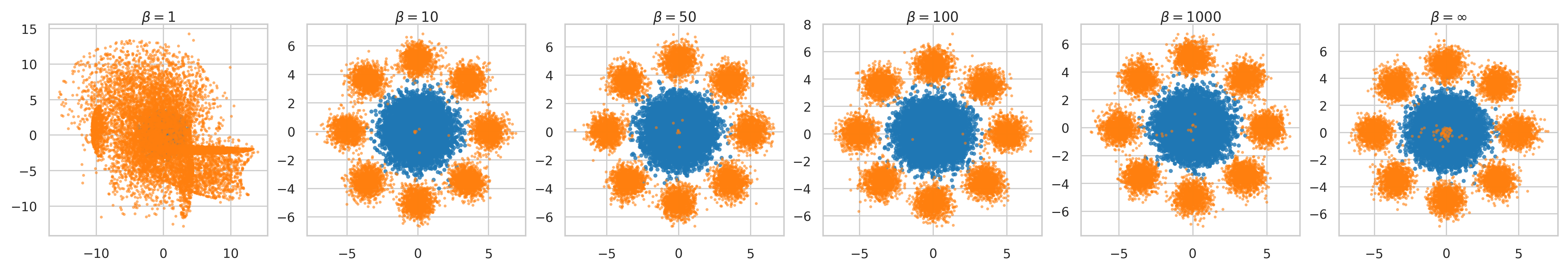}
        \caption{$\mathcal{N}\!\rightarrow\!8\text{gaussians}$}
        \label{fig:n8g}
    \end{subfigure}
    \caption{Transport results across datasets for different $\beta$ values. Initial
    distributions $\mu_0$ are shown in \textit{blue}, and transported distributions
    $\mu_T$ in \textit{orange}.}
    \label{fig:beta_all}
\end{figure}

\paragraph{Influence of $\beta$.}
Figure~\ref{fig:w2_beta} (left) shows that $\mathcal{W}_2$ distances decrease
rapidly as $\beta$ increases from $10$ to $100$, except for
$\mathcal{N}\!\rightarrow\!8\text{gaussians}$ which keeps increasing, with optimal
performance achieved between $\beta = 10$ and $100$ across all datasets. Beyond
$\beta = 100$, performance slightly degrades, suggesting an intermediate $\beta$
value provides the best trade-off between drift and volatility.
Figure~\ref{fig:beta_all} visualizes the transport results for all $\beta$ values.
In each subplot, the initial distribution $\mu_0$ is shown in \textit{blue}, and
the transported distribution $\mu_T$ is shown in \textit{orange}. For the smallest
tested value, $\beta = 1$, the algorithm fails to converge. This behavior can be
attributed to the violation of the required condition $\beta > \frac{1}{T}$, which
is not satisfied when $T = 1$.

\paragraph{Convergence of the outer loop.}
Figure~\ref{fig:w2_beta} (right) shows the normalized 2-Wasserstein distance
$\mathcal{W}_2(k) / \mathcal{W}_2(1)$ as a function of the number of outer
iterations $K$ for two synthetic benchmarks with both $\beta = 100$ and $\beta=10$, in order to illustrate the two regimes. Values are
normalized by the first-iteration distance to allow comparison across tasks with
different absolute scales, and averaged across multiple seeds. Both tasks exhibit rapid convergence within the first
3--5 iterations, after which the $\mathcal{W}_2$ distance plateaus. The
$\mathcal{N} \to \text{8gaussians}$ task converges monotonically, while
$\mathcal{N} \to \text{Moons}$ stabilizes with minor fluctuations but shows no
divergence. This empirically justifies the choice of $K = 5$ used throughout our
experiments and suggests that the alternating procedure converges in practice,
despite the absence of a formal convergence proof for the outer loop.

\paragraph{Practical selection of $\beta$.}
The hyperparameter $\beta$ controls the trade-off between drift and volatility:
larger $\beta$ recovers the classical SB, while smaller $\beta$ increases the role
of stochastic volatility. In practice, one can choose $\beta$ via a coarse grid
search (e.g., $\beta \in \{1, 10, 50, 100\}$) using a held-out metric such as the
2-Wasserstein distance. On the synthetic benchmarks, the optimal $\beta$
consistently lies in $[10, 100]$ (Table~\ref{tab:all_beta_2d}), and performance is
not sensitive to the precise value within this range. The boundary value $\beta T = 1$ exhibits setting-dependent behavior: it fails on the 2D benchmarks (Table~\ref{tab:all_beta_2d}), but yields the best age translation quality on the ALAE latent space (Table~\ref{tab:alae_quant}). By Proposition~\ref{prop:envelope}, the map $\msY_t$ remains well-defined only while $D^2\psi^*$ stays strictly below $\beta I_d$, but as $\beta T \to 1$, this margin vanishes. The 2D targets are sharply multimodal or geometrically non-convex, inducing high-curvature potentials that exhaust this margin, whereas the ALAE latent space is \textit{regularized} by the pretrained autoencoder toward a smooth geometry~\citet{pidhorskyi2020adversarial} that remains within the admissibility bound. When $\beta T$ falls well below $1$ (typically from values around $\beta = 0.6$, $T = 1$), the algorithm fails across all settings, confirming that $\beta T > 1$ remains necessary in practice.

\subsection{Unpaired Image-to-Image Translation Setup}

For unpaired image-to-image translation, we closely follow the experimental setup in \cite{pmlr-v235-gushchin24a}. The dataset consists of $70,000$ labeled images, of which $60,000$ were used for training and $10,000$ for testing. The model hyperparameters were set to $t_{\text{model}}=32$ and $d_{\text{model}}=128$, with $J=10$ potentials in the Gaussian mixture. This results in a drift model with approximately $2.6$ million parameters, and a transport map estimator with $100,000$ parameters. Training took under 5 minutes, and inference on 10 input images required less than 1 minute.  

\begin{figure*}[!t]
    \centering
    \begin{subfigure}[b]{0.48\textwidth}
        \centering
        \includegraphics[width=\textwidth]{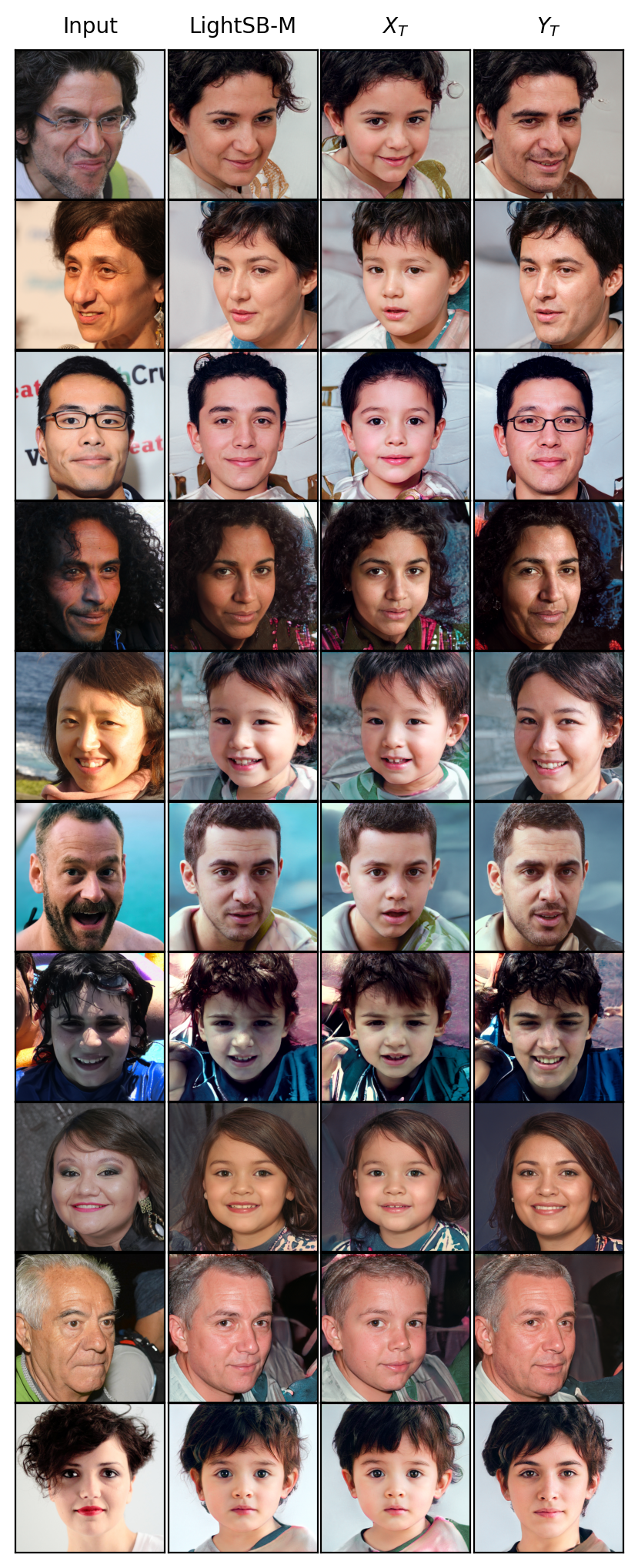}
        \subcaption{$\eps=0.1$}
    \end{subfigure}
    \hfill
    \begin{subfigure}[b]{0.48\textwidth}
        \centering
        \includegraphics[width=\textwidth]{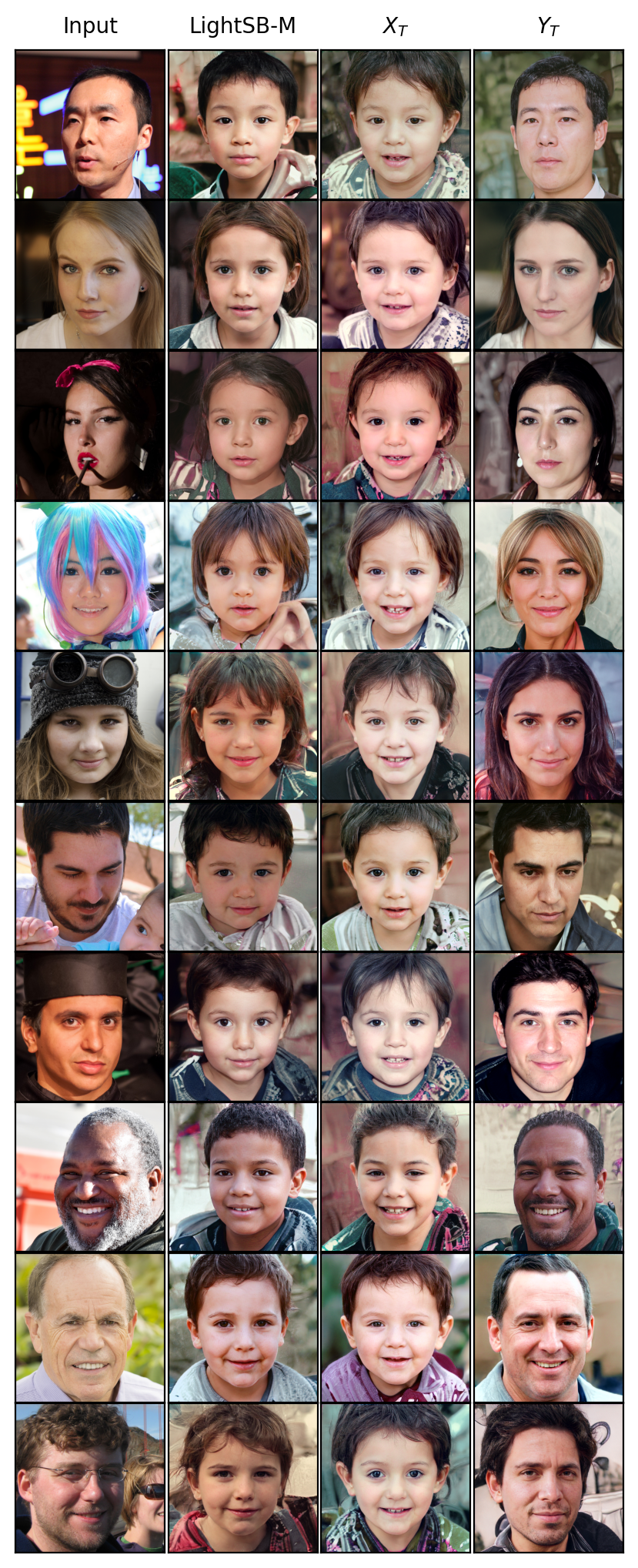}
        \subcaption{$\eps=1$}
    \end{subfigure}
\caption{Additional examples for unpaired image-to-image translation using LightSB-M as the baseline. Results show both $X_T$ and $Y_T$ from SBB for $\beta = 1$ and $\eps \in \{0.1, 1\}$.}

    \label{fig:alae_plots_extra}
\end{figure*}

Our implementation builds upon the code provided by \cite{pmlr-v235-gushchin24a}, available at: \url{https://github.com/SKholkin/LightSB-Matching/tree/main}. Figure~\ref{fig:alae_plots_extra} provides additional examples of our method, showing both $X_T$ and $Y_T$ for $\beta=1$ and $\eps \in \{0.1, 1\}$, with LightSB-M used as the baseline for comparison.

\end{document}